\documentclass[a4paper,fleqn]{cas-dc}    

\usepackage[authoryear,longnamesfirst]{natbib}

\usepackage{amsmath}
\usepackage{amssymb}
\usepackage{mathtools}
\usepackage{multirow}
\usepackage{booktabs}      
\usepackage{xcolor}
\usepackage{colortbl}
\usepackage{dsfont}        
\usepackage{algorithm}     
\usepackage{algorithmic}   
\usepackage[capitalize,noabbrev]{cleveref}  
\usepackage{float}
\usepackage{placeins}
\usepackage{capt-of}
\definecolor{ForestGreen}{rgb}{0.13, 0.55, 0.13}

\graphicspath{ {./figures/} }

\begin{document}
\def\floatpagepagefraction{1}
\def\textpagefraction{.001}

\shorttitle{}
\shortauthors{}

\makeatletter
\let\ps@first\ps@plain
\let\ps@cas\ps@plain
\makeatother

\title[mode=title]{ Domain-Aware Pruning: Sparsity and Domain Generalization via Regularized Probabilistic Masking}

\author[1]{Parham Sazdar}

\ead{p.sazdar@ut.ac.ir}
\credit{Conceptualization, Methodology, Software, Formal analysis, Investigation, Writing -- original draft, Writing -- review \& editing}

\affiliation[1]{organization={Department of Electrical and Computer Engineering, University of Tehran},
                city={Tehran},
                country={Iran}}

\author[1]{Mostafa Tavassolipour}
\cormark[1]
\ead{tavassolipour@ut.ac.ir}
\credit{Methodology, Validation, Writing -- review \& editing}

\author[1]{Reshad Hosseini}
\ead{reshad.hosseini@ut.ac.ir}
\credit{Supervision, Conceptualization, Methodology, Writing -- review \& editing}

\cortext[cor1]{Corresponding author}

\begin{abstract}
Domain generalization (DG) and neural network pruning are conventionally treated as distinct objectives, targeting out-of-distribution (OOD) robustness and model efficiency, respectively. In this work, we bridge this gap by introducing Domain-Aware Pruning (DAP), a framework that leverages network sparsity as a mechanism to implicitly enhance generalization to unseen domains. Diverging from standard binary mask optimization, DAP learns a continuous parameter retention probability $p \in [0, 1]$, framing network compression as a continuous probabilistic masking problem. By introducing a regularization objective that actively penalizes the retention of domain-sensitive weights during the mask training, DAP identifies a domain-invariant subnetwork. Empirical results across five DG benchmark datasets demonstrate that DAP achieves significant sparsity while consistently matching or exceeding the OOD performance of its dense counterparts. Crucially, DAP is an algorithm-agnostic framework that integrates seamlessly with existing DG pipelines without necessitating post-hoc fine-tuning. Beyond efficiency and generalization, we show that DAP natively provides increased robustness to adversarial perturbations and yields highly interpretable models, where the retained weights reliably encapsulate the most domain-invariant and task-critical representations.
\end{abstract}



\maketitle

\pagestyle{plain}
\thispagestyle{plain}

\section{Introduction}

Deep neural networks have achieved remarkable success across a wide range of applications, from computer vision~\citep{he2016deep, dosovitskiy2020image} to natural language processing~\citep{vaswani2017attention}. Yet, their ubiquitous deployment in real-world scenarios faces two fundamental challenges: computational efficiency and generalization under distribution shift. Modern networks often contain millions of parameters, demanding substantial computational resources that limit deployment on resource-constrained edge devices~\citep{han2015learning}. Simultaneously, these models frequently fail when deployed in environments that differ from their training distribution~\citep{gulrajani2020search} a critical limitation for applications ranging from medical diagnosis across different hospitals to autonomous vehicles operating in diverse weather conditions~\citep{koh2021wilds}.

Neural network pruning and domain generalization have emerged as two prominent research directions addressing these challenges independently. Pruning methods remove redundant parameters to create compact and efficient models~\citep{liu2017learning, frankle2019lottery}, while Domain Generalization (DG) techniques aim to learn representations that generalize effectively to unseen target domains~\citep{blanchard2011generalizing, arjovsky2019invariant}. However, these two objectives have been studied largely in isolation. While recent studies have begun to explore pruning in Domain Adaptation (DA) settings~\citep{yu2019transfer, chen2019cooperative}, they typically require access to target domain data. This overlooks a fundamental question in the strict DG setting: \textit{Do all weights contribute equally to generalization, and can we leverage this insight to prune more intelligently without seeing the target domain data?}

We argue that the answer is no and that this insight unlocks a powerful synergy between pruning and generalization. Standard pruning methods, predominantly based on weight magnitude~\citep{han2015learning} or single-domain gradients~\citep{sanh2020movement}, operate under the implicit assumption that larger weights are more important. Yet, this criterion is agnostic to \textit{what} these weights encode. Recent analyses~\citep{long2024rethinking, mayilvahanan2025search} suggest that models often rely on spurious correlations domain-specific artifacts that minimize training loss but harm out-of-distribution performance. In the high-sparsity regime, identifying and removing these "fat" parameters becomes critical~\citep{wang2023recovery}.

Our key insight is that weights exhibiting consistent gradient behavior across multiple source domains are likely encoding domain-invariant features, while weights with conflicting gradients may be capturing domain-specific patterns. By quantifying this through \textbf{Cross-Domain Gradient Alignment}, we can identify which weights to preserve and which to prune—yielding sparse networks that not only match but often exceed the generalization performance of their dense counterparts.

This observation leads to a surprising finding supported by recent theoretical works~\citep{bartoldson2020generalization, zhang2021subnetwork}: sparsity, when applied judiciously, acts as a regularizer for domain generalization. At moderate sparsity levels ($40$--$60\%$), our approach consistently outperforms dense baselines (e.g., ERM~\citep{gulrajani2020search}, MIRO~\citep{cha2022domain}) on out-of-distribution accuracy. This reframes pruning not merely as a compression technique, but as a principled approach to improving generalization by removing domain-specific capacity that would otherwise memorize spurious correlations.

We propose \textbf{DAP}, a framework that integrates domain sensitivity directly into the pruning criterion. Our method maintains a learnable mask over network weights, trained via Gumbel-Softmax relaxation~\citep{jang2016categorical, maddison2016concrete} to enable gradient-based optimization. Crucially, we modulate pruning decisions based on a domain sensitivity score: weights with agreeing gradients (domain-invariant) receive protection from pruning, while weights with conflicting gradients (domain-specific) are preferentially removed. Rather than learning fixed binary masks directly, we learn the \textit{distribution} of masks parameterized by continuous logits. This formulation enables smooth gradient-based optimization and allows the model to express uncertainty about pruning decisions during training, converging to deterministic masks only at inference.
Our main contributions are summarized as follows:
\begin{itemize}
    \item \textbf{Novel Domain-Aware Pruning (DAP) Framework:} We propose a pruning criterion based on cross-domain gradient alignment to identify domain-invariant weights without requiring target domain data, addressing the limitations of prior DA-based pruning~\citep{wang2023recovery}. Our end-to-end approach utilizes differentiable masking~\citep{louizos2017learning} and adaptive scheduling, which eliminates the need for expensive fine-tuning iterations~\citep{han2015learning} and provides fine-grained control to halt training at any desired sparsity level.
    
    \item \textbf{Sparsity as a Domain Generalization Regularizer:} Through extensive experiments on five standard benchmarks (PACS~\citep{li2017deeper}, VLCS~\citep{fang2013unbiased}, TerraIncognita~\citep{beery2018recognition}, OfficeHome~\citep{venkateswara2017deep}, and DomainNet~\citep{peng2019moment}), we demonstrate that DAP achieves $80\%$ sparsity while retaining over $98\%$ of dense model OOD accuracy---significantly outperforming magnitude-based pruning and recent OOD scoring methods~\citep{sun2023pruning, cai2022towards}. Notably, at moderate sparsity ($40\%$), our pruned models exceed dense baseline accuracy, challenging the conventional view of pruning as purely a compression technique.
    
    \item \textbf{Enhanced Robustness and Method Agnosticism:} We empirically show that removing domain-specific weights via DAP effectively filters out vulnerabilities to distribution shifts, thereby improving robustness against natural corruptions. Furthermore, our criterion is base-method agnostic and can be seamlessly integrated as a plug-and-play module into existing state-of-the-art DG algorithms (e.g., ERM, CORAL~\citep{sun2016deep}, MIRO~\citep{cha2022domain}).
\end{itemize}

\section{Background and Related Work}

Our approach lies at the intersection of neural network pruning and domain generalization. Traditionally studied in isolation, recent findings suggest a deep synergy between structural sparsity and out-of-distribution robustness.

\subsection{Neural Network Pruning: From Compression to Regularization}

Network pruning reduces model complexity by removing redundant parameters. Early unstructured methods~\citep{han2015learning} achieved high compression rates but required specialized hardware. Structured pruning, such as channel pruning~\citep{li2017pruning, he2017channel}, removes entire architectural units. A prominent baseline, Network Slimming~\citep{liu2017learning}, utilizes Batch Normalization scaling factors to identify and prune insignificant channels. More advanced iterative methods, like Progressive Channel Pruning~\citep{guo2020progressive}, refine this selection via reconstruction error minimization. A parallel line of work casts pruning as a black-box search problem solved with population-based metaheuristics: \citet{tmamna2024cnn}, for example, formulate filter pruning as a constrained binary particle-swarm optimization over a reduced search space, obtaining compact image classifiers with substantial memory and energy savings.

\textbf{Pruning for Generalization.} Beyond compression, recent theoretical works suggest pruning acts as a regularizer. The \textit{Lottery Ticket Hypothesis} (LTH)~\citep{frankle2019lottery} posits that dense networks contain sparse subnetworks capable of matching full accuracy. \citet{bartoldson2020generalization} formalized the \textit{generalization-stability tradeoff}, showing that pruning induces instability akin to noise injection, guiding optimization toward flatter minima associated with better generalization. Similarly, \citet{zhang2021subnetwork} proposed the ``Functional Lottery Ticket Hypothesis,'' demonstrating that even in biased models, invariant subnetworks exist that outperform the dense network on out-of-distribution (OOD) data. Extending this perspective to a multi-objective setting, \citet{poyatos2023multiobjective} evolve sparse subnetworks with a genetic algorithm that jointly optimizes accuracy, complexity, and robustness---treating robustness as a first-class quality indicator and using the most influential retained neurons to interpret the pruned model.

\textbf{Differentiable Masking.} To enable end-to-end structural learning, recent methods leverage continuous relaxations of discrete variables. \citet{jang2016categorical} and \citet{maddison2016concrete} introduced the Gumbel-Softmax (or Concrete) distribution, allowing gradients to propagate through stochastic nodes. \citet{louizos2017learning} extended this to $L_0$ regularization using Hard-Concrete distributions for exact zero-weight pruning. In contrast to the population-based metaheuristics above, which explore the discrete pruning space without gradient information, differentiable masking optimizes the retention decisions directly by backpropagation, enabling end-to-end integration with representation learning. We build upon these differentiable mechanisms to learn domain-invariant structures directly via gradient descent.

\subsection{Domain Generalization: Benchmarks and Challenges}

Domain Generalization (DG) aims to learn models from source domains $\mathcal{D}_S = \{D_1, \ldots, D_K\}$ that generalize to unseen target domains $D_T$. Standard baselines include Empirical Risk Minimization (ERM) and domain-invariant learning methods like CORAL~\citep{sun2016deep} and GroupDRO~\citep{sagawa2019distributionally}. Recent optimization-based methods like SWAD~\citep{cha2021swad} (seeking flat minima) and MIRO~\citep{cha2022domain} (mutual information regularization) have established state-of-the-art results on the DomainBed benchmark~\citep{gulrajani2020search}.

\textbf{The Generalization Paradox.} A critical challenge in DG is the tradeoff between \textit{discriminability} and \textit{generalizability}. \citet{long2024rethinking} argue that aggressive invariance learning often discards discriminative features, motivating solutions that compress away spurious information while preserving what transfers. Whereas most such solutions operate at the level of learned \textit{representations}, our work pursues a complementary route, inducing analogous compression directly in the \textit{parameter} space through learned sparsity. Furthermore, \citet{mayilvahanan2025search} recently demonstrated that the perceived success of large-scale foundation models on DG tasks is often an ``illusion'' caused by data contamination, and true OOD generalization remains an open challenge requiring structural solutions rather than just more data.

\subsection{Pruning Under Distribution Shift}

Pruning methods designed for i.i.d. data often fail under distribution shift. We categorize existing cross-domain pruning works into two groups:

\textbf{1. Pruning for Domain Adaptation (Target-Dependent).}
Most research focuses on Domain Adaptation (DA), where target data is available. \citet{yu2019transfer} proposed Transfer Channel Pruning (TCP) using target-aware MMD loss. \citet{chen2019cooperative} used cooperative learning between source and target models. More recently, \citet{yang2024domain} introduced Domain Adaptive Channel Pruning (DACP) for edge deployment. \citet{wang2023recovery} theoretically showed that standard pruning increases domain discrepancy, necessitating target-specific fine-tuning.
\textit{Limitation:} These methods fundamentally rely on target data access, making them inapplicable to DG scenarios.

\textbf{2. Pruning for Domain Generalization (Target-Free).}
Research in pruning specifically for DG is scarce. \citet{sanh2020movement} introduced \textit{Movement Pruning} for transfer learning, showing that preserving weights moving away from zero (gradient-based) is superior to preserving large weights (magnitude-based) under shift. In DG, \citet{cai2022towards} proposed an importance score based on the variance of domain-level risks. \citet{sun2023pruning} utilized a domain similarity score.
\textit{Limitation:} These methods are often ``reactive'' (relying on output loss variance) or apply pruning as a post-processing step. They lack an end-to-end mechanism to proactively align gradients during the training of the mask.

We address this gap with DAP. Unlike DA methods, DAP is target-free. Unlike prior DG pruning, DAP is end-to-end and ``proactive,'' leveraging \textit{Cross-Domain Gradient Alignment} to identify and preserve invariant mechanisms before the loss variance diverges.

\section{Domain Aware Pruning}
\label{sec:method}

We present \textbf{Domain-Aware Pruning (DAP)}, a framework that integrates domain sensitivity into neural network sparsification. By moving beyond magnitude-based criteria, DAP identifies and preserves parameters that exhibit invariant behavior across domains. We first formalize the problem setup, then describe our learnable pruning mechanism based on differentiable relaxation, introduce the domain sensitivity score derived from gradient alignment, and finally detail the training objective with adaptive coefficient scheduling.

\subsection{Problem Setup}

We consider the Domain Generalization (DG) setting with $K$ source domains $\mathcal{D}_S = \{D_1, D_2, \ldots, D_K\}$, where each domain $D_k = \{(x_i^k, y_i^k)\}_{i=1}^{n_k}$ contains input-label pairs drawn from a domain-specific distribution $P_k(X, Y)$. Our goal is to learn a sparse neural network $f_\theta: \mathcal{X} \rightarrow \mathcal{Y}$ parameterized by $\theta$ that generalizes to an unseen target domain $D_T$ drawn from a different distribution $P_T(X, Y)$, where $P_T \neq P_k$.

Let $\mathbf{W} = \{\mathbf{W}^{(l)}\}_{l=1}^{L}$ denote the weight tensors across $L$ layers. We seek a binary mask $\mathbf{M} = \{\mathbf{M}^{(l)}\}_{l=1}^{L}$, where $\mathbf{M}^{(l)} \in \{0, 1\}^{|\mathbf{W}^{(l)}|}$, such that the masked network $f_{\mathbf{W} \odot \mathbf{M}}$ achieves:
\begin{itemize}
    \item \textbf{High sparsity:} $\frac{1}{|\theta|}\sum_{l} \|\mathbf{M}^{(l)}\|_0 \leq 1 - s$, where $s$ is the target sparsity ratio.
    \item \textbf{Strong generalization:} Minimized expected risk on the unseen target domain:\\
    $\mathbb{E}_{(x,y) \sim P_T}[\ell(f_{\mathbf{W} \odot \mathbf{M}}(x), y)]$.
\end{itemize}
Here, $|\theta| = \sum_{l} |\mathbf{W}^{(l)}|$ denotes the total number of parameters in the dense network, and $\ell(\cdot, \cdot)$ is the task loss function (e.g., cross-entropy).


\subsection{Learnable Pruning Masks via Continuous Relaxation}

Direct optimization over binary masks is intractable due to the discrete nature of $\mathbf{M}^{(l)} \in \{0, 1\}^{|\mathbf{W}^{(l)}|}$. To enable end-to-end learning via gradient descent, we adopt the continuous relaxation approach commonly used in differentiable architecture search and $L_0$ regularization~\citep{louizos2017learning}.
For each weight tensor $\mathbf{W}^{(l)}$, we introduce a corresponding logit tensor $\mathbf{\Lambda}^{(l)} \in \mathbb{R}^{|\mathbf{W}^{(l)}|}$ of the same shape. Following the parameterization of the Hard-Concrete distribution~\citep{louizos2017learning}, the probability of keeping weight $w_{i}^{(l)}$ is modeled as:
\begin{equation}
p_{i}^{(l)} = \sigma(\lambda_{i}^{(l)}) = \frac{1}{1 + \exp(-\lambda_{i}^{(l)})},
\end{equation}
where $\sigma(\cdot)$ is the sigmoid function and $\lambda_{i}^{(l)}$ is the $i$-th element of the learnable logit tensor $\boldsymbol{\Lambda}^{(l)}$.
DAP operates at the level of individual weights 
(\textit{unstructured pruning}): each scalar parameter 
$w_{i}^{(l)}$ has its own mask $m_{i}^{(l)}$. This provides maximum 
flexibility in identifying domain-specific parameters 
at arbitrary locations in the network, at the cost of 
requiring sparse tensor support for inference acceleration.
 A key distinction of our approach is that we learn the \textit{distribution} over binary masks rather than fixed masks. Each logit $\lambda_{i}^{(l)}$ parameterizes a Bernoulli distribution over the keep/prune decision. This probabilistic formulation offers significant advantages: (1) it enables smooth optimization over the discrete mask space; (2) it allows the model to capture uncertainty about weight importance during training; and (3) it provides controllable sparsity, as the distribution can be sampled or thresholded at any training stage to yield a valid sparse network.

Direct sampling from a Bernoulli distribution is a non-differentiable operation, which blocks gradient flow during backpropagation. To enable end-to-end optimization of the mask distribution, we employ the Gumbel-Softmax (also known as the Concrete distribution)~\citep{jang2016categorical, maddison2016concrete} as a continuous, differentiable approximation of discrete sampling via the reparameterization trick.

We use the Straight-Through (ST) estimator: in the forward pass, the continuous Gumbel-Softmax output is used directly without discretization to preserve differentiability; in the backward pass, gradients flow through the continuous softmax approximation. Concretely, for each weight $w_{i}^{(l)}$, the mask value is computed as:
\begin{equation}
m_{i}^{(l)} = \text{ST-Gumbel-Softmax}\left(\begin{bmatrix} 0 \\ \lambda_{i}^{(l)} \end{bmatrix}, \tau\right),
\end{equation}
where the input is a two-logit vector whose first entry is fixed at $0$ (corresponding to the ``prune'' class) and whose second entry is $\lambda_{i}^{(l)}$ (corresponding to the ``keep'' class), and $\tau > 0$ is the temperature parameter governing the sharpness of the approximation.

The temperature $\tau$ is annealed exponentially over $T$ total training steps:
\begin{equation}
\tau_t = \tau_{\text{init}} \cdot \left(\frac{\tau_{\text{final}}}{\tau_{\text{init}}}\right)^{t/T},
\end{equation}
where $t$ is the current training step, and $\tau_{\text{final}} < \tau_{\text{init}}$. We begin with a high temperature $\tau_{\text{init}}$, under which the Gumbel-Softmax output approaches a uniform distribution, yielding smoother gradients that facilitate early exploration. As training progresses, $\tau$ is reduced toward $\tau_{\text{final}}$, causing the output to concentrate on near-discrete values and approximate a true Bernoulli sample though at the cost of higher gradient variance. This schedule balances optimization stability in early training with near-discrete mask decisions at convergence.

\paragraph{Inference.} At test time, we apply deterministic hard masks based on the learned logits:
\begin{equation}
m_{i}^{(l)} = \mathds{1}[\lambda_{i}^{(l)} > 0].
\end{equation}
It is worth noting that since $\sigma(\lambda_{i}^{(l)}) > 0.5 \iff \lambda_{i}^{(l)} > 0$, this is equivalent to pruning all weights whose learned keep probability falls below $0.5$. This thresholding yields a sparse network that can be efficiently deployed. The continuous relaxation during training allows the model to express uncertainty about which weights to prune, while the final hard thresholding ensures a deterministic, interpretable mask at inference.

\subsection{Domain Sensitivity via Gradient Alignment}
Before delving into the technical details, it is worth establishing 
the core intuition behind our approach. Our primary motivation is to 
develop a mechanism in which $\lambda$ acts as a dynamic guide for 
each individual weight. Specifically, we assign a score to each 
weight reflecting how well it captures invariant representations 
across domains: weights that consistently encode domain-invariant 
features receive high scores and are preferentially retained, while 
weights that remain domain-specific are assigned low scores and 
suppressed. Crucially, by embedding this logic directly into the 
behavior of $\lambda$ and its associated scoring system, we achieve 
domain generalization without the need to introduce additional 
complex terms into the loss function.


Standard pruning methods rely on weight magnitude~\citep{han2015learning} 
or single-domain gradients~\citep{sanh2020movement}. However, in DG, a 
weight with high magnitude might encode a domain-specific artifact 
rather than a generalizable feature. We introduce a \textbf{Domain 
Sensitivity Score} based on gradient alignment to explicitly quantify 
the domain-invariance of each weight.

Consider a weight $w$ in the network. 
During training on multiple source domains, each domain provides 
a gradient signal indicating how $w$ should be updated to minimize 
that domain's loss. If gradients from all domains agree in sign, 
the weight is learning a domain-invariant feature; if they 
disagree in sign, it is being pulled toward domain-specific 
solutions. This observation aligns with a common strategy in 
domain generalization: identifying and suppressing network 
components that encode domain-specific patterns, thereby 
encouraging the model to rely on transferable representations.

For each source domain $D_k$, we 
compute the gradient of the base training objective with respect to 
the weights:
\begin{equation}
\mathbf{g}_k^{(l)} = \nabla_{\mathbf{W}^{(l)}} \mathcal{L}_{\text{base}}(D_k; \theta),
\end{equation}

where $\mathcal{L}_{\text{base}}$ is the loss function used to 
pretrain the backbone  e.g., the cross-entropy loss 
$\mathcal{L}_{\text{CE}}$ for ERM, or the full training objective 
(including regularization terms) for methods such as 
CORAL~\citep{sun2016deep} or MIRO~\citep{cha2022domain}. This 
ensures that the domain sensitivity score reflects the gradient 
geometry of the actual model being pruned.

To quantify how aligned the gradients are across domains at the level of 
individual weights, we use sign agreement --- a lightweight, 
magnitude-invariant measure of directional consistency. For each 
weight $w$, we compare the signs of its gradients across domain 
pairs:
\begin{equation}
a_{mn}[w] = \text{sign}(g_m[w]) \cdot \text{sign}(g_n[w]).
\end{equation}
This element-wise product equals $+1$ when both domains $m$ and $n$ 
agree on the update direction (both positive or both negative), and 
$-1$ when they disagree. Note that sign agreement can be viewed as 
element-wise cosine similarity where each component is normalized 
to unit magnitude, preserving only directional information while 
being substantially cheaper to compute.

We aggregate the alignment 
scores over all $\binom{K}{2}$ domain pairs:
\begin{equation}
S[w] = -\frac{2}{K(K-1)} \sum_{m=1}^{K} \sum_{n=m+1}^{K} a_{mn}[w],
\end{equation}
where $m$ and $n$ are domain indices. Given $K$ source domains, this 
score is computed for all $\binom{K}{2}$ pairwise combinations (e.g., 
for $K=3$: pairs $(1,2)$, $(1,3)$, and $(2,3)$), and the result is 
averaged to yield a single domain sensitivity score per weight.

The negation ensures that the score has an intuitive interpretation:
\begin{itemize}
    \item $S[w] \to -1$: High alignment across all domains 
    $\Rightarrow$ \textbf{Domain-Invariant} $\Rightarrow$ Preserve
    \item $S[w] \to +1$: Low alignment (conflicting gradients) 
    $\Rightarrow$ \textbf{Domain-Sensitive} $\Rightarrow$ Prune
\end{itemize}

Gradient alignment computed on 
mini-batches can be noisy. To obtain stable estimates, we maintain 
an Exponential Moving Average (EMA) of the sensitivity score:
\begin{equation}
\bar{S}_t[w^{(l)}_{i}] = \beta \cdot \bar{S}_{t-1}[w^{(l)}_{i}] + (1 - \beta) \cdot \bar{S}_t[w^{(l)}_{i}],
\end{equation}
where $\beta \in [0, 1)$ controls the smoothing strength. Higher 
$\beta$ yields more stable but slower-adapting estimates.
Computing sign agreement 
requires only $K$ backward passes (one per domain) every 
$f_{\text{update}}$ steps. With typical settings ($K=3$, 
$f_{\text{update}}=100$), this adds negligible overhead compared to the standard training loop, especially since the gradients can be computed in parallel across domains.

\subsection{Domain-Aware Pruning Criterion}

We integrate domain sensitivity directly into the learnable masking process by modifying the logits used for sampling.

The pruning probability is determined by the \textit{effective logits}:
\begin{equation}\label{eq:lambda_update}
\tilde{\lambda}_{i}^{(l)} = \lambda_{i}^{(l)} - \alpha \cdot \bar{S}[w^{(l)}_{i}],
\end{equation}
where $\alpha \geq 0$ controls the influence of domain sensitivity.

This formulation creates a principled "push-pull" mechanism:
\begin{itemize}
    \item For \textbf{domain-invariant weights} ($\bar{S} < 0$): The subtraction of a negative value \textit{increases} the effective logit, raising the keep probability.
    \item For \textbf{domain-sensitive weights} ($\bar{S} > 0$): The subtraction of a positive value \textit{decreases} the effective logit, raising the prune probability.
\end{itemize}

When $\alpha = 0$, our method reduces to domain-agnostic pruning based solely on the learned logits, providing a natural ablation baseline.
\subsection{Training Objective}

The total objective function balances task performance and 
structural sparsity:
\begin{equation}
\mathcal{L}_{\text{total}} = \mathcal{L}_{\text{base}} + \lambda_s \cdot \mathcal{L}_{\text{sparsity}}.
\end{equation}

We minimize the average empirical risk 
over source domains. Since DAP operates on an ERM-trained 
backbone, the task loss is the standard cross-entropy:
\begin{equation}
\mathcal{L}_{\text{base}} = \frac{1}{K} \sum_{k=1}^{K} \mathbb{E}_{(x,y) \sim D_k} \left[ -\log f_{\mathbf{W} \odot \mathbf{M}}(x)[y] \right].
\end{equation}
When DAP is applied to a different base method (e.g., CORAL 
or MIRO), $\mathcal{L}_{\text{base}}$ is replaced by the 
corresponding training objective, consistent with the gradient 
computation in Equation~(5).

To enforce the target sparsity 
ratio $s$, we minimize the squared deviation between expected 
density and target density:
\begin{equation}
\mathcal{L}_{\text{sparsity}} = \left( \frac{1}{|\theta|} \sum_{l} \sum_{i} \sigma(\tilde{\lambda}_{i}^{(l)}) - (1 - s) \right)^2.
\end{equation}
This symmetric penalty discourages both under-pruning and 
over-pruning, guiding the model toward exactly the desired 
sparsity level.

\subsection{Adaptive Coefficient Scheduling}
A fixed sparsity coefficient $\lambda_s$ often leads to training instabilities: too small delays convergence to target sparsity; too large causes premature pruning and accuracy collapse. We propose an adaptive scheduling mechanism:
\begin{equation}
\lambda_s^{(t)} = \lambda_{\text{base}} \cdot \phi_{\text{progress}}^{(t)} \cdot \phi_{\text{difficulty}}^{(t)} \cdot \phi_{\text{ratio}}^{(t)},
\end{equation}
where $\lambda_{\text{base}}$ is a fixed base coefficient (a hyperparameter whose value is reported in Appendix~\ref{appendix:hyperparameters}), which sets the overall scale of the sparsity penalty and is then dynamically modulated by the three time-dependent factors $\phi_{\text{progress}}^{(t)}$, $\phi_{\text{difficulty}}^{(t)}$, and $\phi_{\text{ratio}}^{(t)}$, each of which adapts the coefficient in response to the current state of training.

 Gradually increases pressure as sparsity approaches the target:
\begin{equation}
\phi_{\text{progress}}^{(t)} = 1 + a \cdot \frac{s_t}{s_{\text{target}}},
\end{equation}
where $s_t$ is the current sparsity and $a$ is an acceleration constant.

Accounts for the increasing difficulty of pruning at high sparsity levels, where each remaining weight becomes more critical:
\begin{equation}
\phi_{\text{difficulty}}^{(t)} = \frac{1}{(1 - s_t + \epsilon)^{\rho}} - 1,
\end{equation}
where $\rho$ controls the exponent and $\epsilon$ prevents numerical instability.

Dynamically balances the magnitudes of the task and sparsity losses:
\begin{equation}
\phi_{\text{ratio}}^{(t)} = \text{clip}\left(\frac{\mathcal{L}_{\text{base}}}{\mathcal{L}_{\text{sparsity}} + \epsilon} \cdot \frac{1}{r_{\text{target}}}, \, r_{\min}, \, r_{\max}\right),
\end{equation}
where $\mathcal{L}_{\text{base}}$ denotes the loss with which the backbone was originally pretrained. For example, cross-entropy under ERM, or the full MIRO objective if the backbone was pretrained with MIRO ensuring that the ratio factor remains consistent with the base training method. $r_{\text{target}}$ is the desired loss ratio, and clipping prevents extreme values. The quantities $r_{\text{target}}$, $r_{\min}$, and $r_{\max}$ are treated as hyperparameters and are calibrated empirically via manual tuning on source-domain validation performance; their chosen values for each benchmark are reported in Appendix~\ref{appendix:hyperparameters}.

 The coefficient is smoothed via EMA to prevent oscillations:
\begin{equation}
\bar{\lambda}_s^{(t)} = \eta \cdot \bar{\lambda}_s^{(t-1)} + (1 - \eta) \cdot \lambda_s^{(t)}.
\end{equation}

The complete DAP procedure is summarized in Algorithm~\ref{alg:dap}.

\begin{algorithm}[t]
\caption{Domain-Aware Pruning (DAP)}
\label{alg:dap}
\small
\begin{algorithmic}[1]
\REQUIRE Domains $\{D_1, \ldots, D_K\}$, weights $\mathbf{W}$,
         sparsity $s$, steps $T$
\ENSURE Sparse mask $\mathbf{M}$
\STATE Init logits $\mathbf{\Lambda}^{(l)}$ s.t.\ 
       $\sigma(\mathbf{\Lambda}) \approx p_{\text{init}}$
\STATE Freeze weights $\mathbf{W}$;\ \ 
       $\bar{S}[w]^{(l)} \leftarrow \mathbf{0}$
\FOR{$t = 1$ to $T$}
    \STATE Sample $\{(X_k, Y_k)\}_{k=1}^K$ from each domain
    \STATE $\tau_t \leftarrow \tau_{\text{init}} \cdot 
           (\tau_{\text{final}}/\tau_{\text{init}})^{t/T}$
    \IF{$t \bmod f_{\text{upd}} = 0$ \AND $\alpha > 0$}
        \FOR{$k = 1$ to $K$}
            \STATE $\mathbf{g}_k^{(l)} \leftarrow 
                   \nabla_{\mathbf{W}^{(l)}} 
                   \mathcal{L}_{\text{base}}(D_k)$
        \ENDFOR
        \STATE $a_{mn} \leftarrow 
               \text{sign}(\mathbf{g}_m) \odot 
               \text{sign}(\mathbf{g}_n)$
        \STATE $S[w^{(l)}] \leftarrow 
               \tfrac{-2}{K(K{-}1)} \sum_{m<n} a_{mn}$
        \STATE $\bar{S}[w^{(l)}] \leftarrow 
               \beta\, \bar{S}[w^{(l)}] + (1{-}\beta)\, S[w^{(l)}]$
    \ENDIF
    \STATE $\tilde{\mathbf{\Lambda}}^{(l)} \leftarrow 
           \mathbf{\Lambda}^{(l)} - \alpha \cdot \bar{S}[w^{(l)}]$
    \STATE $\mathbf{M}^{(l)} \sim 
           \text{GumbelST}(\tilde{\mathbf{\Lambda}}^{(l)}, \tau_t)$
    \STATE $\hat{Y} \leftarrow f_{\mathbf{W} \odot \mathbf{M}}(X)$
    \STATE Compute $\mathcal{L}_{\text{CE}}$, 
           $\mathcal{L}_{\text{sp}}$;\ \ 
           update $\bar{\lambda}_s^{(t)}$ via Eq.\,(17)
    \STATE $\mathcal{L} \leftarrow \mathcal{L}_{\text{CE}} + 
           \bar{\lambda}_s^{(t)}\, \mathcal{L}_{\text{sp}}$
    \STATE $\mathbf{\Lambda} \leftarrow \mathbf{\Lambda} - 
           \eta_\Lambda \nabla_{\mathbf{\Lambda}} \mathcal{L}$
\ENDFOR

\STATE $\mathbf{M}^{(l)} \leftarrow 
 \mathds{1}[\tilde{\lambda}_{i}^{(l)} > 0]$
\STATE \textbf{return} $\mathbf{M}$
\end{algorithmic}
\end{algorithm}
\subsection{Practical Considerations}

\paragraph{Controllable Sparsity.} A key advantage of our framework is that training can be stopped at any point to obtain a deployable sparse model. Since masks are determined by thresholding logits, the model at step $t$ is immediately usable at its current sparsity level $s_t$ without additional post-processing.



\paragraph{No Fine-tuning Required.} Unlike iterative pruning methods that require retraining after each pruning step~\citep{han2015learning, frankle2019lottery}, our approach produces high-quality sparse networks directly. The pretrained weights remain frozen throughout; only the mask logits are optimized. If desired, optional fine-tuning can be applied after pruning by unfreezing weights and training with a small learning rate.

\paragraph{Method Agnosticism.} While we present DAP using ERM as the base training objective, the domain sensitivity score is agnostic to the underlying method. DAP can be applied as a plug-and-play module to any domain generalization algorithm (e.g., CORAL~\citep{sun2016deep}, DANN~\citep{ganin2016domain}, SWAD~\citep{cha2021swad}) by computing gradient alignment with respect to that method's training loss.

\paragraph{Interpretability.} Beyond compression, our framework provides interpretability by quantifying the importance of individual weights for cross-domain generalization. The domain sensitivity scores reveal which parameters encode transferable knowledge versus domain-specific patterns, offering insights into what the network has learned.

\section{Experiments}
We conduct a comprehensive empirical evaluation of the proposed method (DAP) 
to validate its core claims and situate it within the broader landscape of 
domain generalization and model compression research. Our evaluation is 
structured around the following research questions:

\begin{itemize}
    \item \textbf{RQ1 — Generalization under sparsity:} Can sparse networks 
    produced by DAP match or exceed the out-of-distribution accuracy of dense 
    domain generalization baselines, and is there an optimal sparsity level at 
    which performance peaks?

    \item \textbf{RQ2 — Training stability:} Does mask learning remain stable 
    throughout training? We examine whether accuracy degrades during the 
    pruning process and how the sparsity-accuracy trade-off evolves across 
    training epochs.

    \item \textbf{RQ3 — Algorithmic agnosticism:} Is DAP limited to ERM-trained 
    models, or does it provide consistent gains when applied to models trained 
    with other domain generalization objectives?

    \item \textbf{RQ4 — Pruning criterion quality:} Does domain-aware weight 
    selection outperform domain-agnostic pruning baselines? We compare DAP 
    against magnitude pruning, Taylor pruning, random pruning, and recent 
    methods that operate under distribution shift.

    \item \textbf{RQ5 — Robustness:} Do the sparse subnetworks identified by 
    DAP exhibit improved robustness to natural corruptions and adversarial 
    perturbations relative to their dense counterparts?

    \item \textbf{RQ6 — Interpretability:} What does the learned logit 
    distribution reveal about which weights the network relies on for 
    cross-domain generalization?
\end{itemize}

\paragraph{Experimental Setup.}
All experiments use a ResNet-50~\citep{he2016deep} backbone pretrained on 
ImageNet. We evaluate on five standard benchmarks from the DomainBed 
suite~\citep{gulrajani2020search}: \textbf{PACS}~\citep{li2017deeper} (4 
domains, 7 classes), \textbf{VLCS}~\citep{fang2013unbiased} (4 domains, 5 
classes), \textbf{OfficeHome}~\citep{venkateswara2017deep} (4 domains, 65 
classes), \textbf{TerraIncognita}~\citep{beery2018recognition} (4 domains, 
10 classes), and \textbf{DomainNet}~\citep{peng2019moment} (6 domains, 345 
classes). These datasets span a wide range of domain shift types, scales, 
and class granularities, providing a rigorous test bed for evaluating 
generalization under sparsity. Following the DomainBed protocol~\citep{gulrajani2020search}, 
we report leave-one-domain-out accuracy averaged across all 
test environments. Each environment is evaluated with 3 
independent seeds; $\pm$ values denote standard deviations 
across seeds.

For each dataset, we first train a dense ERM model following the standard 
DomainBed protocol, then apply DAP to learn binary masks over the frozen 
pretrained weights. Crucially, the original weights are never modified 
during mask learning.  DAP identifies a functional sparse subnetwork 
within the pretrained backbone without any fine-tuning. We adopt the 
training-domain validation model selection criterion 
from~\citet{gulrajani2020search} as our primary protocol. Hyperparameters 
are tuned independently per dataset using source-domain validation 
performance. Dense ERM baselines are from our own training runs using the 
\href{https://github.com/facebookresearch/DomainBed}{DomainBed codebase} to ensure a matched comparison with DAP, 
which prunes the same checkpoint. Results for all other dense 
methods are taken directly from~\citet{gulrajani2020search}.
Note that all sparsity levels reported for DAP are extracted as 
checkpoints from a single training run targeting 99.9\% sparsity, 
rather than from independent runs optimized for each level. The 20\% and 40\% models are therefore early snapshots along the path to the final target. Our results are thus a conservative lower bound on DAP's potential at each sparsity level; dedicated per-target training would likely yield further gains.

DAP freezes all pretrained weights and batch normalization 
statistics, training only the mask logits $\lambda$ via Adam.
Temperature anneals exponentially from $\tau = 2.0$ 
(exploration) to $\tau = 0.3$ (near-deterministic) over 
$T = 100{,}000$ steps. The adaptive coefficient $\lambda_s$ 
follows a feedback-driven schedule combining progress, 
difficulty, and loss-ratio signals with EMA smoothing. 
The domain sensitivity strength $\alpha$ and learning rate 
$\eta$ are tuned per dataset; all other hyperparameters are 
shared (Appendix~\ref{appendix:hyperparameters}).

\subsection{Comparison with Domain Generalization Methods}
\label{sec:main_results}

We address \textbf{RQ1}: whether sparse networks produced by DAP 
can match or surpass dense domain generalization methods. 
Table~\ref{tab:main_results} reports OOD accuracy for DAP at four 
sparsity levels across five DomainBed benchmarks, alongside eleven 
dense baselines including state-of-the-art methods MIRO~\citep{cha2022domain} 
and SWAD~\citep{cha2021swad}.

\begin{table*}[t]
\centering
\caption{
    \textbf{Comparison with domain generalization methods.} 
    Out-of-distribution accuracy (\%) on five DomainBed benchmarks 
    (ResNet-50 backbone). DAP achieves accuracy comparable to the 
    strongest dense methods while using only 20--40\% of parameters. 
    Best in \textbf{bold}, second best \underline{underlined}. 
    $^\dagger$Sparse model. 
    All results averaged over 3 seeds; $\pm$ denotes standard deviation across seeds.
}
\label{tab:main_results}
\small
\resizebox{0.85\textwidth}{!}{%
\begin{tabular}{lccccccc}
\toprule
\textbf{Method} & \textbf{Sparsity} & \textbf{PACS} & \textbf{VLCS} & \textbf{OfficeHome} & \textbf{TerraInc} & \textbf{DomainNet} & \textbf{Avg} \\
\midrule
\multicolumn{8}{l}{\textit{Dense Models (100\% parameters)}} \\
\midrule
ERM             & 0\% & 85.7\;{\tiny$\pm$0.5} & 77.4\;{\tiny$\pm$0.3} & 67.7\;{\tiny$\pm$0.5} & \textbf{54.2\;{\tiny$\pm$0.4}} & 43.6\;{\tiny$\pm$0.2} & 65.7 \\
IRM             & 0\% & 84.4\;{\tiny$\pm$1.1} & 78.1\;{\tiny$\pm$0.0} & 66.6\;{\tiny$\pm$1.0} & 47.9\;{\tiny$\pm$0.7} & 35.7\;{\tiny$\pm$1.9} & 62.5 \\
GroupDRO        & 0\% & 84.1\;{\tiny$\pm$0.4} & 77.2\;{\tiny$\pm$0.6} & 66.9\;{\tiny$\pm$0.3} & 47.0\;{\tiny$\pm$0.3} & 33.7\;{\tiny$\pm$0.2} & 61.8 \\
Mixup           & 0\% & 84.3\;{\tiny$\pm$0.5} & 77.7\;{\tiny$\pm$0.4} & 69.0\;{\tiny$\pm$0.1} & 48.9\;{\tiny$\pm$0.8} & 39.6\;{\tiny$\pm$0.1} & 63.9 \\
MLDG            & 0\% & 84.8\;{\tiny$\pm$0.6} & 77.1\;{\tiny$\pm$0.4} & 68.2\;{\tiny$\pm$0.1} & 46.1\;{\tiny$\pm$0.8} & 41.8\;{\tiny$\pm$0.4} & 63.6 \\
CORAL           & 0\% & 86.0\;{\tiny$\pm$0.2} & 77.7\;{\tiny$\pm$0.5} & 68.6\;{\tiny$\pm$0.4} & 46.4\;{\tiny$\pm$0.8} & 41.8\;{\tiny$\pm$0.2} & 64.1 \\
MMD             & 0\% & 85.0\;{\tiny$\pm$0.2} & 76.7\;{\tiny$\pm$0.9} & 67.7\;{\tiny$\pm$0.1} & 49.3\;{\tiny$\pm$1.4} & 39.4\;{\tiny$\pm$0.8} & 63.6 \\
DANN            & 0\% & 84.6\;{\tiny$\pm$1.1} & 78.7\;{\tiny$\pm$0.3} & 65.4\;{\tiny$\pm$0.6} & 48.4\;{\tiny$\pm$0.5} & 38.4\;{\tiny$\pm$0.0} & 63.1 \\
C-DANN          & 0\% & 82.8\;{\tiny$\pm$1.5} & 78.2\;{\tiny$\pm$0.4} & 65.6\;{\tiny$\pm$0.5} & 47.6\;{\tiny$\pm$0.8} & 38.9\;{\tiny$\pm$0.1} & 62.6 \\
MIRO            & 0\% & 85.4\;{\tiny$\pm$0.4} & \underline{79.0\;{\tiny$\pm$0.0}} & \underline{70.5\;{\tiny$\pm$0.4}} & 50.4\;{\tiny$\pm$1.1} & 44.3\;{\tiny$\pm$0.2} & \underline{65.9} \\
SWAD            & 0\% & \textbf{88.1\;{\tiny$\pm$0.1}} & \textbf{79.1\;{\tiny$\pm$0.1}} & \textbf{70.6\;{\tiny$\pm$0.2}} & 50.0\;{\tiny$\pm$0.3} & \textbf{46.5\;{\tiny$\pm$0.1}} & \textbf{66.9} \\
\midrule
\multicolumn{8}{l}{\textit{Sparse Models (DAP, Ours)}} \\
\midrule
DAP$^\dagger$ & 20\% & 85.8\;{\tiny$\pm$0.1} & 77.2\;{\tiny$\pm$0.1} & 69.0\;{\tiny$\pm$0.2} & \underline{53.2\;{\tiny$\pm$0.2}} & \underline{44.5\;{\tiny$\pm$0.5}} & \underline{65.9} \\
DAP$^\dagger$ & 40\% & \underline{87.8\;{\tiny$\pm$0.1}} & 76.8\;{\tiny$\pm$0.2} & 69.9\;{\tiny$\pm$0.2} & 51.1\;{\tiny$\pm$0.2} & 43.9\;{\tiny$\pm$0.2} & \underline{65.9} \\
DAP$^\dagger$ & 60\% & 87.2\;{\tiny$\pm$0.2} & 75.1\;{\tiny$\pm$0.4} & 65.9\;{\tiny$\pm$0.5} & 48.2\;{\tiny$\pm$0.5} & 42.6\;{\tiny$\pm$0.3} & 63.8 \\
DAP$^\dagger$ & 80\% & 86.0\;{\tiny$\pm$0.4} & 72.8\;{\tiny$\pm$0.7} & 62.3\;{\tiny$\pm$0.4} & 46.4\;{\tiny$\pm$0.9} & 43.5\;{\tiny$\pm$0.4} & 62.2 \\
\bottomrule
\end{tabular}%
}
\end{table*}

At 20--40\% sparsity (i.e., 20--40\% of the network's parameters 
have been pruned away), DAP matches or surpasses the majority of 
dense baselines  including its own ERM backbone  while 
using 20--40\% fewer parameters. On average, DAP at this 
operating point trails only SWAD among all eleven methods 
evaluated, and matches MIRO exactly. The strongest individual 
result is on TerraIncognita, where DAP at 20\% sparsity is the 
top-performing method overall, exceeding every dense baseline 
including MIRO and SWAD. On PACS, DAP at 40\% sparsity 
approaches the best dense result. These findings confirm that 
dense models allocate substantial capacity to domain-specific 
features that harm generalization, and that their selective 
removal via gradient alignment acts as an effective implicit 
regularizer. Importantly, DAP is orthogonal to training-time 
methods: since it operates post-hoc on frozen weights, it can 
be applied to models trained with any algorithm 
(Section~\ref{sec:plug_and_play}).

Beyond the optimal sparsity range, accuracy degrades gradually 
rather than catastrophically. At 60\% sparsity, DAP remains 
competitive with the dense ERM baseline on average using only 
40\% of parameters; at 80\%, retaining one in five weights, it 
still surpasses several dense methods on individual benchmarks. 
The rate of degradation is dataset-dependent, reflecting the 
severity of the underlying distribution shift. PACS and DomainNet 
are remarkably resilient to aggressive pruning --- and in fact 
benefit from it, as the removal of domain-specific weights acts 
as an effective regularizer that improves OOD generalization. 
TerraIncognita, by contrast, exhibits a steeper decline: its 
camera-trap images span geographically distinct locations with 
pronounced visual shift, leaving the network with little 
redundant capacity. In this regime, each retained weight carries 
greater marginal value for generalization, making pruning 
decisions considerably more sensitive and the cost of removing 
an invariant weight considerably higher. A practitioner can select any point 
along the sparsity spectrum and obtain a deployable model 
without retraining.

\subsection{Training Dynamics and Sparsity-Accuracy Trade-off}
\label{sec:training_dynamics}

\begin{figure*}[!ht]
    \centering
    \includegraphics[width=\textwidth]{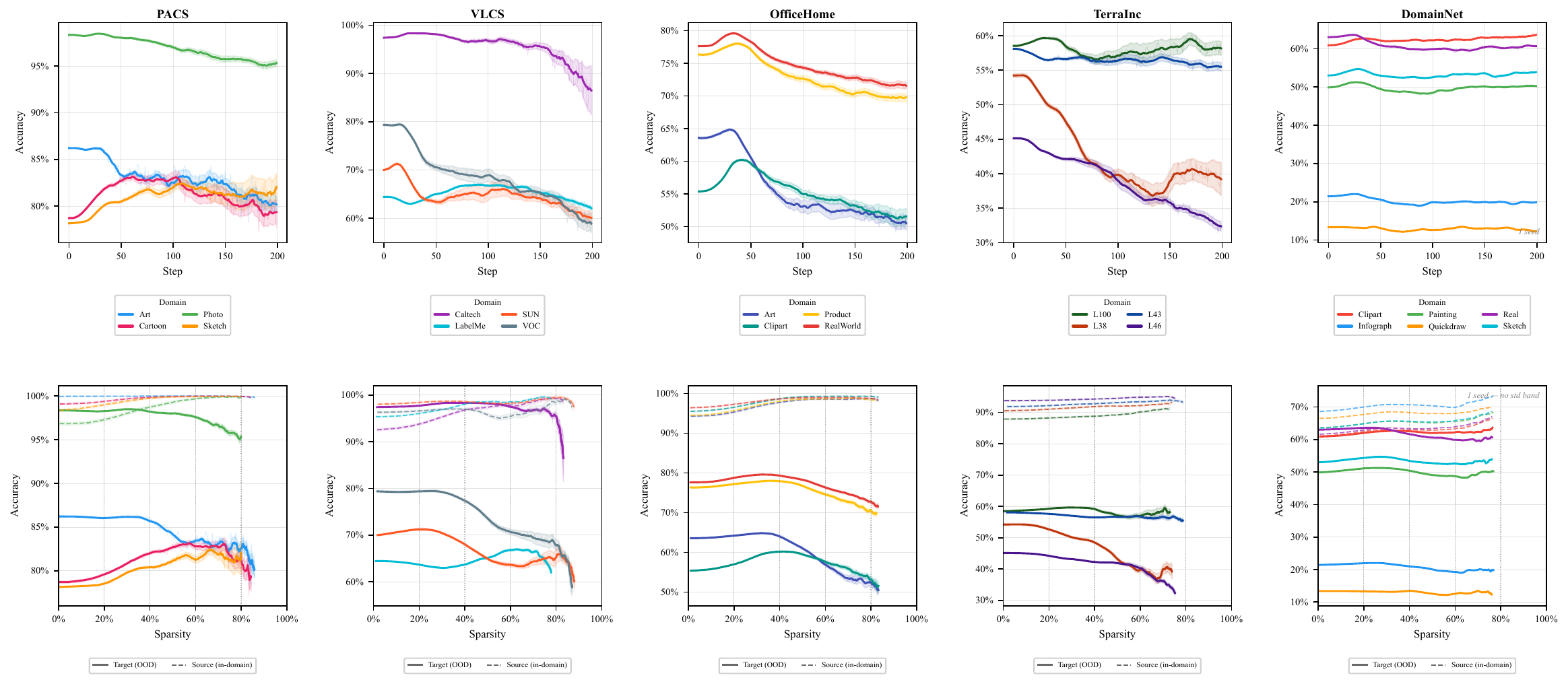}
    \caption{
        \textbf{Training dynamics and sparsity-accuracy trade-offx
        across all five DomainBed benchmarks.}
        \textit{Top row:} target-domain accuracy over training steps
        (smoothed, averaged over 3 seeds, $\pm$1 std).
        \textit{Bottom row:} accuracy vs.\ sparsity ratio; solid lines
        denote target-domain (OOD) accuracy and dashed lines denote
        source (in-domain) accuracy.
        The Cartoon domain of PACS (top row, leftmost column,
        \textcolor{pink}{pink}) is a clear representative example:
        accuracy rises to a peak around step~75--100, then declines
        gracefully as sparsity pressure increases — with no
        catastrophic drops or oscillations.
        In the bottom row, OOD accuracy peaks at ${\sim}$40--60\%
        sparsity and \textit{exceeds} the dense baseline before
        declining at higher compression, while source accuracy remains
        near-perfect throughout — the empirical signature of
        sparsity-as-regularization.
    }
    \label{fig:training_dynamics_combined}
\end{figure*}

We address \textbf{RQ2}: whether mask learning is stable throughout
training and how OOD generalization evolves across the sparsity
spectrum. Figure~\ref{fig:training_dynamics_combined} summarises
both dimensions across all five DomainBed benchmarks.

The non-monotonic pattern visible across all datasets confirms that
pruning selectively removes domain-specific capacity rather than core
representations. At moderate sparsity, spurious correlations are
eliminated; only at extreme sparsity does the criterion begin
discarding genuinely invariant weights. The optimal range
consistently falls at 40--60\%, aligning with
Table~\ref{tab:main_results}.

\subsection{Algorithmic Agnosticism: DAP as a Plug-and-Play Module}
\label{sec:plug_and_play}

We address \textbf{RQ3}: whether DAP's benefits extend beyond 
ERM to models trained with other DG objectives. We apply DAP to 
two base methods representing complementary philosophies: 
\textbf{CORAL}~\citep{sun2016deep}, which aligns feature 
distributions by modifying weights during training, and 
\textbf{MIRO}~\citep{cha2022domain}, which regularizes 
representations to stay close to the pretrained backbone. 
Table~\ref{tab:plug_and_play} reports results on PACS and 
OfficeHome.

\begin{table}[!tb]
\centering
\caption{
    \textbf{Plug-and-play compatibility of DAP.} OOD accuracy 
    (\%) when DAP is applied on top of CORAL and MIRO pretrained 
    models on PACS and OfficeHome. $\Delta$ denotes improvement 
    over the respective dense baseline. $^\dagger$Sparse model.
}
\label{tab:plug_and_play}
\resizebox{\columnwidth}{!}{%
\begin{tabular}{llccc}
\toprule
\textbf{Dataset} 
    & \textbf{Method} 
    & \textbf{Sparsity} 
    & \textbf{Avg} 
    & \textbf{$\Delta$ $\uparrow$} \\
\midrule

\multirow{10}{*}{PACS}
    & CORAL (Dense)         & 0\%  
        & 84.3 $\pm$ 0.0 &  \\
    & CORAL + DAP$^\dagger$ & 20\% 
        & 84.4 $\pm$ 0.1 
        & \textcolor{ForestGreen}{+0.1} \\
    & CORAL + DAP$^\dagger$ & 40\% 
        & 85.2 $\pm$ 0.2 
        & \textcolor{ForestGreen}{+0.9} \\
    & CORAL + DAP$^\dagger$ & 60\% 
        & 86.0 $\pm$ 0.4 
        & \textcolor{ForestGreen}{+1.7} \\
    & CORAL + DAP$^\dagger$ & 80\% 
        & \textbf{86.9 $\pm$ 0.4} 
        & \textcolor{ForestGreen}{\textbf{+2.6}} \\
\cmidrule{2-5}
    & MIRO (Dense)          & 0\%  
        & 87.8 $\pm$ 0.0 &  \\
    & MIRO + DAP$^\dagger$  & 20\% 
        & 88.0 $\pm$ 0.1 
        & \textcolor{ForestGreen}{+0.2} \\
    & MIRO + DAP$^\dagger$  & 40\% 
        & 88.4 $\pm$ 0.2 
        & \textcolor{ForestGreen}{+0.6} \\
    & MIRO + DAP$^\dagger$  & 60\% 
        & \textbf{89.1 $\pm$ 0.3} 
        & \textcolor{ForestGreen}{\textbf{+1.3}} \\
    & MIRO + DAP$^\dagger$  & 80\% 
        & 87.9 $\pm$ 0.5 
        & \textcolor{ForestGreen}{+0.1} \\

\midrule

\multirow{10}{*}{OfficeHome}
    & CORAL (Dense)         & 0\%  
        & 65.4 $\pm$ 0.0 &  \\
    & CORAL + DAP$^\dagger$ & 20\% 
        & 66.5 $\pm$ 0.1 
        & \textcolor{ForestGreen}{+1.1} \\
    & CORAL + DAP$^\dagger$ & 40\% 
        & 68.4 $\pm$ 0.1 
        & \textcolor{ForestGreen}{+3.0} \\
    & CORAL + DAP$^\dagger$ & 60\% 
        & \textbf{68.7 $\pm$ 0.3} 
        & \textcolor{ForestGreen}{\textbf{+3.3}} \\
    & CORAL + DAP$^\dagger$ & 80\% 
        & 67.0 $\pm$ 0.3 
        & \textcolor{ForestGreen}{+1.6} \\
\cmidrule{2-5}
    & MIRO (Dense)          & 0\%  
        & 71.7 $\pm$ 0.1 &  \\
    & MIRO + DAP$^\dagger$  & 20\% 
        & \textbf{72.1 $\pm$ 0.2} 
        & \textcolor{ForestGreen}{\textbf{+0.4}} \\
    & MIRO + DAP$^\dagger$  & 40\% 
        & 72.0 $\pm$ 0.2 
        & \textcolor{ForestGreen}{+0.3} \\
    & MIRO + DAP$^\dagger$  & 60\% 
        & 70.0 $\pm$ 0.4 
        & \textcolor{red}{$-$1.7} \\
    & MIRO + DAP$^\dagger$  & 80\% 
        & 66.1 $\pm$ 0.4 
        & \textcolor{red}{$-$5.6} \\
\bottomrule
\end{tabular}%
}
\end{table}

DAP improves both base methods at moderate sparsity.
DAP consistently improves over the dense baseline for both 
base methods at moderate sparsity levels, confirming its 
plug-and-play character. The $\Delta$ column in 
Table~\ref{tab:plug_and_play} shows positive gains across 
both datasets and both base methods up to 60\% sparsity, 
with the largest improvements appearing on CORAL. These 
results confirm that domain-specific weights persist even in 
models trained with explicit invariance objectives, and that 
DAP successfully identifies and removes them post-hoc.

Larger gains on CORAL than MIRO reveal complementarity.
The gains from DAP are consistently larger on top of CORAL 
than MIRO. CORAL aligns feature distributions by modifying 
weight \textit{values} but does not address which 
\textit{connections} encode domain-specific behavior.  DAP 
answers precisely this structural question, making the two 
approaches highly complementary. MIRO, by contrast, 
regularizes representations to stay close to the pretrained 
backbone, partially accomplishing what DAP does post-hoc: 
suppressing domain-specific weight updates during training. 
The MIRO backbone therefore contains fewer domain-sensitive 
weights for DAP to remove, yielding smaller marginal gains 
and earlier onset of over-pruning on the more complex 
OfficeHome benchmark. Practitioners should apply DAP to 
MIRO-trained models at moderate sparsity on complex datasets, 
where gains are consistent and over-pruning risk is minimal.

\subsection{Comparison with Pruning Methods}
\label{sec:pruning_comparison}


We address \textbf{RQ4}: does domain-aware weight selection 
outperform domain-agnostic pruning baselines? All methods are 
applied to the same pretrained ERM backbone without fine-tuning, 
unless otherwise noted. We compare against \textbf{Random Pruning}, 
\textbf{Magnitude Pruning}~\citep{han2015learning}, and 
\textbf{Taylor Pruning}~\citep{molchanov2019importance}, as well 
as two methods designed for pruning under distribution shift: 
\textbf{IoR}~\citep{cai2022towards} (Taylor-based importance 
augmented with the variance of domain-level risks) and 
\textbf{DSS}~\citep{sun2023pruning} (channel scoring via 
cross-domain activation similarity). Both IoR and DSS require 
fine-tuning after pruning and employ structured (channel-level) 
sparsity. Results are in Table~\ref{tab:pruning_comparison}.

\begin{table}[!thb]
\centering
\caption{
    \textbf{Comparison with pruning baselines.} OOD accuracy 
    (\%) on PACS and OfficeHome at multiple sparsity levels. 
    All methods applied to pretrained ERM backbone. 
    $^\dagger$Requires fine-tuning after pruning. 
    `---' denotes unreported or inapplicable.
    \textbf{Bold}: best sparse result per sparsity level.
}
\label{tab:pruning_comparison}
\resizebox{\columnwidth}{!}{%
\begin{tabular}{llcccc}
\toprule
\multirow{2}{*}{\textbf{Method}} 
    & \multirow{2}{*}{\textbf{Sparsity}} 
    & \multicolumn{2}{c}{\textbf{PACS}} 
    & \multicolumn{2}{c}{\textbf{OfficeHome}} \\
\cmidrule(lr){3-4} \cmidrule(lr){5-6}
    & & \textbf{Avg} & \textbf{$\Delta$ $\uparrow$} 
      & \textbf{Avg} & \textbf{$\Delta$ $\uparrow$} \\
\midrule
Dense (ERM) & 0\% & 84.38 & --- & 67.67 & --- \\
\midrule
\multicolumn{6}{l}{\textit{Domain-Agnostic Pruning (No Fine-tuning)}} \\
\midrule
Random 
    & 20\% & 19.96 & \textcolor{red}{$-$64.4} 
           & 2.16  & \textcolor{red}{$-$65.5} \\
Random 
    & 40\% & 22.02 & \textcolor{red}{$-$62.4} 
           & 0.98  & \textcolor{red}{$-$66.7} \\
Random 
    & 60\% & 13.68 & \textcolor{red}{$-$70.7} 
           & 0.99  & \textcolor{red}{$-$66.7} \\
Random 
    & 80\% & 10.34 & \textcolor{red}{$-$74.0} 
           & 1.49  & \textcolor{red}{$-$66.2} \\
\midrule
Magnitude~\cite{han2015learning} 
    & 20\% & 84.23 & \textcolor{red}{$-$0.2} 
           & 67.62 & \textcolor{red}{$-$0.1} \\
Magnitude~\cite{han2015learning} 
    & 40\% & 84.32 & \textcolor{red}{$-$0.1} 
           & 66.98 & \textcolor{red}{$-$0.7} \\
Magnitude~\cite{han2015learning} 
    & 60\% & 82.31 & \textcolor{red}{$-$2.1} 
           & 63.24 & \textcolor{red}{$-$4.4} \\
Magnitude~\cite{han2015learning} 
    & 80\% & 46.92 & \textcolor{red}{$-$37.5} 
           & 17.77 & \textcolor{red}{$-$49.9} \\
\midrule
Taylor~\cite{molchanov2017pruning} 
    & 20\% & 79.72 & \textcolor{red}{$-$4.7} 
           & 65.06 & \textcolor{red}{$-$2.6} \\
Taylor~\cite{molchanov2017pruning} 
    & 40\% & 72.69 & \textcolor{red}{$-$11.7} 
           & 58.18 & \textcolor{red}{$-$9.5} \\
Taylor~\cite{molchanov2017pruning} 
    & 60\% & 32.54 & \textcolor{red}{$-$51.8} 
           & 17.82 & \textcolor{red}{$-$49.9} \\
Taylor~\cite{molchanov2017pruning} 
    & 80\% & 16.52 & \textcolor{red}{$-$67.9} 
           & 2.41  & \textcolor{red}{$-$65.3} \\
\midrule
\multicolumn{6}{l}{\textit{Domain-Aware Pruning (Prior Work, Fine-tuning Required)}} \\
\midrule
IoR$^{\dagger}$~\cite{cai2022towards} (structured) 
    & 50\% & 84.54 & \textcolor{ForestGreen}{$+$0.2} 
           & ---   & --- \\
IoR$^{\dagger}$~\cite{cai2022towards} (structured) 
    & 70\% & 80.68 & \textcolor{red}{$-$3.7} 
           & ---   & --- \\
DSS$^{\dagger}$~\cite{sun2023pruning} (MIRO backbone) 
    & 10\% & 86.50 & \textcolor{ForestGreen}{$+$2.1} 
           & 71.40 & \textcolor{ForestGreen}{$+$3.7} \\
\midrule
\multicolumn{6}{l}{\textit{Domain-Aware Pruning (Ours, No Fine-tuning)}} \\
\midrule
\rowcolor{gray!12}
DAP (Ours) 
    & 20\% & \textbf{85.80} 
           & \textcolor{ForestGreen}{\textbf{$+$1.4}} 
           & \textbf{69.00} 
           & \textcolor{ForestGreen}{\textbf{$+$1.3}} \\
\rowcolor{gray!12}
DAP (Ours) 
    & 40\% & \textbf{87.80} 
           & \textcolor{ForestGreen}{\textbf{$+$3.4}} 
           & \textbf{69.90} 
           & \textcolor{ForestGreen}{\textbf{$+$2.2}} \\
\rowcolor{gray!12}
DAP (Ours) 
    & 60\% & \textbf{87.20} 
           & \textcolor{ForestGreen}{\textbf{$+$2.8}} 
           & \textbf{65.90} 
           & \textcolor{red}{$-$1.8} \\
\rowcolor{gray!12}
DAP (Ours) 
    & 80\% & \textbf{86.00} 
           & \textcolor{ForestGreen}{\textbf{$+$1.6}} 
           & \textbf{62.30} 
           & \textcolor{red}{$-$5.4} \\
\bottomrule
\end{tabular}%
}
\end{table}

\textbf{Domain-agnostic methods collapse; DAP does not.}
The central finding is stark: all three domain-agnostic baselines 
degrade catastrophically at high sparsity, while DAP remains 
functional and even exceeds the dense baseline on PACS. Random 
Pruning is non-functional at every level. Magnitude Pruning 
survives at low sparsity but collapses at 80\%, and Taylor 
Pruning degrades even more rapidly. DAP, by contrast, produces 
deployable models across the full sparsity spectrum. This 
divergence occurs because magnitude and Taylor criteria are blind 
to the distributional role of each weight.

\textbf{Comparison with prior domain-aware methods.}
IoR~\citep{cai2022towards} and DSS~\citep{sun2023pruning} are the 
closest prior works. IoR augments the standard Taylor importance 
score with the variance of domain-level risks; DSS scores 
channels by the similarity of their activation maps across source 
domains. Both operate under more favorable conditions than DAP: 
each requires a fine-tuning stage after pruning to recover 
accuracy, and both employ structured (channel-level) sparsity 
--- which preserves dense tensor operations but limits the 
granularity of weight selection. Additionally, DSS is applied to 
a MIRO-pretrained backbone, a stronger starting point than the 
ERM model DAP prunes. Despite these advantages, neither method 
scales to the high-sparsity regime where pruning decisions become 
critical: IoR is evaluated at 50\% and 70\% filter sparsity on 
PACS only, while DSS is reported at just 10\% channel sparsity 
on DomainBed. DAP, starting from the weaker ERM backbone and 
without any fine-tuning, produces competitive results at low 
sparsity and continues to improve through 40--60\% --- a regime 
where neither IoR nor DSS has been evaluated. The key distinction 
is methodological: IoR and DSS compute importance scores in a 
single pass over a fixed model and then prune, whereas DAP 
\textit{jointly learns} mask logits and domain sensitivity in an 
end-to-end optimization loop, allowing the pruning criterion to 
adapt as the mask evolves.


In the above experiment, we compare against post-hoc pruning methods applied to a fixed 
pretrained model. Iterative methods such as 
LTH~\citep{frankle2019lottery}, which interleave pruning with full 
retraining, address a different problem setting: they modify 
weight values across multiple rounds and are not directly 
comparable to DAP's single-pass, weight-frozen extraction.

\subsection{Robustness to Adversarial Attacks}
\label{sec:robustness}

While the primary objective of DAP is to enhance 
out-of-distribution generalization, we further investigate 
whether the unstructural sparsity it induces carries an 
additional benefit: improved robustness to adversarial 
perturbations. We address \textbf{RQ5} by evaluating 
FGSM~\citep{goodfellow2014explaining} and 
PGD~\citep{madry2017towards} attacks at five perturbation 
severity levels with $\epsilon \in \{0.5, 1.0, 2.0, 4.0, 
8.0\}/255$, corresponding to very weak through strong 
perturbations. For FGSM, only the perturbation budget 
$\epsilon$ varies across levels. For PGD, both the budget 
and the number of steps are scaled with severity: levels 1--2 
($\epsilon \leq 1/255$) use 5 steps and levels 3--5 
($\epsilon \geq 2/255$) use 10 steps, with per-step size 
$\alpha$ set to half the perturbation budget at each level. 
We compare the dense ERM baseline against the 80\%-sparse 
DAP model on both source and target (OOD) data. Results are 
reported in Tables~\ref{tab:robustness_pacs} 
and~\ref{tab:robustness_officehome}.

\begin{table}[!tbh]
\centering
\caption{
    \textbf{Adversarial robustness on PACS.} Accuracy (\%) of 
    Dense vs.\ 80\%-Sparse DAP under FGSM and PGD attacks at 
    five severity levels, on source (Src) and target/OOD (Tgt) 
    data. $\Delta$ = Sparse $-$ Dense. Higher $\Delta$ indicates 
    greater robustness advantage of the sparse model.
}
\label{tab:robustness_pacs}
\resizebox{\columnwidth}{!}{%
\begin{tabular}{llcccccc}
\toprule
\multirow{2}{*}{\textbf{Attack}} 
    & \multirow{2}{*}{\textbf{Sev.}} 
    & \multicolumn{2}{c}{\textbf{Dense}} 
    & \multicolumn{2}{c}{\textbf{Sparse (80\%)}} 
    & \multicolumn{2}{c}{\textbf{$\Delta$ $\uparrow$}} \\
\cmidrule(lr){3-4} \cmidrule(lr){5-6} \cmidrule(lr){7-8}
    & & \textbf{Src} & \textbf{Tgt} 
      & \textbf{Src} & \textbf{Tgt} 
      & \textbf{Src} & \textbf{Tgt} \\
\midrule
Clean & --- 
    & 95.98 & 84.37 
    & 96.19 & 85.20 
    & \textcolor{ForestGreen}{$+$0.21} 
    & \textcolor{ForestGreen}{$+$0.83} \\
\midrule
\multirow{5}{*}{FGSM}
    & 1 & 91.69 & 76.77 & 94.48 & 81.64 
        & \textcolor{ForestGreen}{$+$2.80} 
        & \textcolor{ForestGreen}{$+$4.87} \\
    & 2 & 87.02 & 70.04 & 92.46 & 77.93 
        & \textcolor{ForestGreen}{$+$5.43} 
        & \textcolor{ForestGreen}{$+$7.89} \\
    & 3 & 80.49 & 62.74 & 88.37 & 70.93 
        & \textcolor{ForestGreen}{$+$7.88} 
        & \textcolor{ForestGreen}{$+$8.19} \\
    & 4 & 72.51 & 53.45 & 81.62 & 61.47 
        & \textcolor{ForestGreen}{$+$9.11} 
        & \textcolor{ForestGreen}{$+$8.02} \\
    & 5 & 65.12 & 46.36 & 71.45 & 50.04 
        & \textcolor{ForestGreen}{$+$6.33} 
        & \textcolor{ForestGreen}{$+$3.67} \\
\midrule
\multirow{5}{*}{PGD}
    & 1 & 81.04 & 61.67 & 91.37 & 75.87 
        & \textcolor{ForestGreen}{$+$10.33} 
        & \textcolor{ForestGreen}{$+$14.20} \\
    & 2 & 64.98 & 46.12 & 84.98 & 65.00 
        & \textcolor{ForestGreen}{$+$20.00} 
        & \textcolor{ForestGreen}{$+$18.87} \\
    & 3 & 35.52 & 24.38 & 69.32 & 46.24 
        & \textcolor{ForestGreen}{$+$33.81} 
        & \textcolor{ForestGreen}{$+$21.86} \\
    & 4 & 15.67 & 11.69 & 40.67 & 21.85 
        & \textcolor{ForestGreen}{$+$25.00} 
        & \textcolor{ForestGreen}{$+$10.17} \\
    & 5 & 5.66  & 4.44  & 12.41 & 4.06  
        & \textcolor{ForestGreen}{$+$6.76} 
        & \textcolor{red}{$-$0.39} \\
\bottomrule
\end{tabular}%
}
\end{table}

\begin{table}[!tb]
\centering
\caption{
    \textbf{Adversarial robustness on OfficeHome.} Same 
    protocol as Table~\ref{tab:robustness_pacs}.
}
\label{tab:robustness_officehome}
\resizebox{\columnwidth}{!}{%
\begin{tabular}{llcccccc}
\toprule
\multirow{2}{*}{\textbf{Attack}} 
    & \multirow{2}{*}{\textbf{Sev.}} 
    & \multicolumn{2}{c}{\textbf{Dense}} 
    & \multicolumn{2}{c}{\textbf{Sparse (80\%)}} 
    & \multicolumn{2}{c}{\textbf{$\Delta$ $\uparrow$}} \\
\cmidrule(lr){3-4} \cmidrule(lr){5-6} \cmidrule(lr){7-8}
    & & \textbf{Src} & \textbf{Tgt} 
      & \textbf{Src} & \textbf{Tgt} 
      & \textbf{Src} & \textbf{Tgt} \\
\midrule
Clean & --- 
    & 82.03 & 67.29 
    & 78.33 & 62.21 
    & \textcolor{red}{$-$3.70} 
    & \textcolor{red}{$-$5.08} \\
\midrule
\multirow{5}{*}{FGSM}
    & 1 & 73.08 & 56.13 & 73.29 & 55.83 
        & \textcolor{ForestGreen}{$+$0.22} 
        & \textcolor{red}{$-$0.30} \\
    & 2 & 65.54 & 48.54 & 69.12 & 50.75 
        & \textcolor{ForestGreen}{$+$3.58} 
        & \textcolor{ForestGreen}{$+$2.21} \\
    & 3 & 57.01 & 40.08 & 62.37 & 43.43 
        & \textcolor{ForestGreen}{$+$5.35} 
        & \textcolor{ForestGreen}{$+$3.34} \\
    & 4 & 49.47 & 32.57 & 53.14 & 34.50 
        & \textcolor{ForestGreen}{$+$3.66} 
        & \textcolor{ForestGreen}{$+$1.93} \\
    & 5 & 44.39 & 28.80 & 44.67 & 27.10 
        & \textcolor{ForestGreen}{$+$0.29} 
        & \textcolor{red}{$-$1.70} \\
\midrule
\multirow{5}{*}{PGD}
    & 1 & 56.20 & 39.30 & 66.26 & 47.60 
        & \textcolor{ForestGreen}{$+$10.06} 
        & \textcolor{ForestGreen}{$+$8.31} \\
    & 2 & 39.20 & 24.84 & 54.62 & 36.07 
        & \textcolor{ForestGreen}{$+$15.42} 
        & \textcolor{ForestGreen}{$+$11.22} \\
    & 3 & 19.11 & 11.60 & 34.49 & 19.79 
        & \textcolor{ForestGreen}{$+$15.39} 
        & \textcolor{ForestGreen}{$+$8.19} \\
    & 4 & 8.02  & 5.25  & 15.22 & 8.05  
        & \textcolor{ForestGreen}{$+$7.20} 
        & \textcolor{ForestGreen}{$+$2.80} \\
    & 5 & 2.54  & 1.92  & 4.13  & 2.31  
        & \textcolor{ForestGreen}{$+$1.59} 
        & \textcolor{ForestGreen}{$+$0.39} \\
\bottomrule
\end{tabular}%
}
\end{table}

\textbf{Sparse networks are substantially more robust.}
The 80\%-sparse DAP model outperforms the dense baseline under 
adversarial attack at nearly every severity level on both 
datasets. The advantage is most pronounced under PGD, which 
provides a stronger test of true vulnerability: as the $\Delta$ 
columns show, the robustness gap \textit{grows} with attack 
strength, peaking at mid-severity before both models saturate 
at extreme perturbation budgets. This pattern holds on both 
PACS and OfficeHome, despite the sparse model starting from 
slightly lower clean accuracy on the latter. The widening gap 
indicates that the sparse model is intrinsically more resistant 
to gradient-based perturbations, not simply better-initialized.


We hypothesize that this robustness improvement arises from 
a shared mechanism: the domain-specific weights that DAP 
removes are those most responsive to distributional variation, 
and such weights may also correspond to locally sensitive 
directions that gradient-based adversarial attacks exploit. 
By eliminating these parameters, the sparse network presents 
a smoother loss landscape with fewer exploitable directions. 
This interpretation is consistent with the theoretical link 
between flat loss landscapes and adversarial 
robustness~\citep{madry2017towards}, though we note that 
domain sensitivity (gradient disagreement across domains) and 
adversarial vulnerability (gradient magnitude with respect to 
input perturbations) are distinct quantities. A rigorous 
theoretical characterization of their relationship is an 
interesting direction for future work.

\subsection{Interpretability: Logit Distribution Analysis}
\label{sec:interpretability}

We address \textbf{RQ6}: what does the learned logit distribution 
reveal about weight importance? Unlike magnitude-based criteria, 
DAP learns a continuous logit $\lambda_i$ for every weight, where 
$\sigma(\lambda_i)$ gives the keep probability. 
Figure~\ref{fig:logit_distribution} shows the logit histogram after 
mask learning at 80\% sparsity on PACS (Cartoon domain).


\vspace{1em}
\noindent\begin{minipage}{\columnwidth}
    \centering
    \includegraphics[width=\columnwidth]{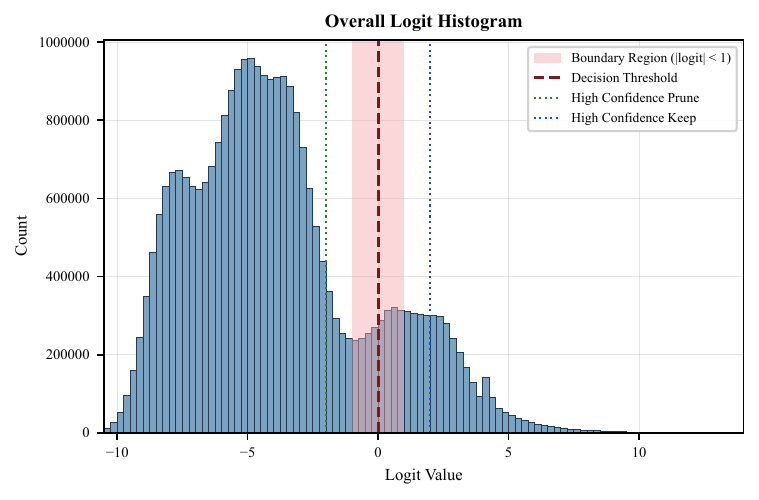}
    \captionof{figure}{\textbf{Logit histogram after mask learning (PACS, Cartoon domain, 80\% sparsity).} The pink shaded region ($|\lambda| < 1$) marks the \textit{boundary region} of uncertain weights. The red dashed line at $\lambda = 0$ is the decision threshold.}
    \label{fig:logit_distribution}
\end{minipage}
\vspace{1em}

The majority of weights 
are pushed to large negative logits ($\lambda \approx -3$ to 
$-10$, where $\sigma(\lambda) < 0.047$), indicating confident 
elimination. A smaller mass occupies large positive logits 
($\lambda > 3$, $\sigma(\lambda) > 0.95$), representing 
weights retained with near-certainty. The boundary region 
($|\lambda| < 1$, $0.27<\sigma(\lambda)<0.73$ ) contains a small fraction of the total 
mass, confirming that the Gumbel-Softmax relaxation and 
adaptive scheduler drive logits toward decisive values. This means DAP produces not only a sparse mask but also 
a continuous importance ranking grounded in cross-domain 
gradient agreement  qualitatively different from 
magnitude-based scores, as a small-magnitude weight with 
$\lambda \gg 0$ is one that all source domains agree should 
be active, while a large-magnitude weight with $\lambda \ll 0$ 
encodes domain-specific behavior despite its size.

\subsection{Ablation Study: The Role of Domain Sensitivity}
\label{sec:ablation}

To We isolate the domain sensitivity score, we set $\alpha = 0$
in Equation~\eqref{eq:lambda_update}, which reduce DAP to a domain-agnostic learnable
pruning method. The Gumbel-Softmax mechanism, adaptive scheduler,
and sparsity objective all remain intact, but the effective logits are no
longer modulated by gradient alignment. Any difference is
therefore attributable solely to the domain sensitivity criterion.

\begin{figure*}[!ht]
    \centering
    \includegraphics[width=\textwidth]{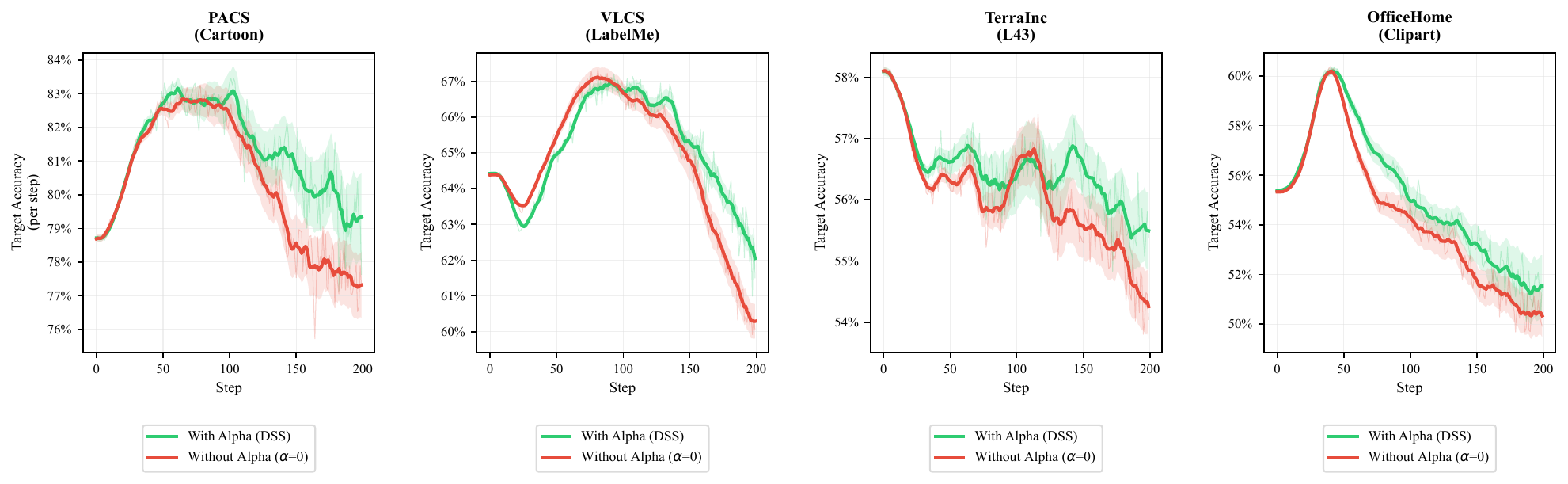}
    \caption{
        \textbf{Ablation: Domain Sensitivity Score (DSS) across
        four benchmarks.}
        Green: full DAP ($\alpha > 0$).
        Red: domain-agnostic variant ($\alpha = 0$).
        Mean OOD accuracy over 3 seeds ($\pm$1 std).
        Columns correspond to representative target domains:
        PACS/Cartoon, VLCS/LabelMe, TerraIncognita/L43, and
        OfficeHome/Clipart.
    }
    \label{fig:ablation_alpha}
\end{figure*}

Figure~\ref{fig:ablation_alpha} compares OOD accuracy
trajectories with ($\alpha > 0$) and without ($\alpha = 0$)
domain sensitivity across four benchmarks and representative
target domains, averaged over 3 seeds.

\noindent\textbf{Domain sensitivity is unnecessary at low
sparsity but critical at high sparsity.}
As Figure~\ref{fig:ablation_alpha} shows, both variants are
nearly indistinguishable during the early training phase when
sparsity is low: with a generous parameter budget, even a
domain-agnostic criterion avoids discarding critical invariant
structure. Beyond this regime the curves diverge and
consistently across all four benchmarks. This progressive
divergence reveals that the domain sensitivity score's value is
\textit{conditional on pruning difficulty}: at high sparsity,
the candidate set shifts from clearly redundant weights to
\textit{contested} weights of comparable learned importance, and
gradient alignment serves as the decisive signal for
distinguishing transferable from domain-specific structure.

\section{Conclusion}
\label{sec:conclusion}

We presented DAP, a framework that
bridges neural network sparsification and domain generalization
through a key insight: not all weights contribute equally to
generalization. By leveraging cross-domain gradient alignment to
preferentially preserve domain-invariant weights, DAP produces sparse
networks that are not only efficient but also more robust and
generalizable.

Experiments across five DomainBed benchmarks show that sparsity, when
applied judiciously, acts as a regularizer --- pruned models
\textit{exceed} dense baseline accuracy at moderate sparsity levels,
challenging the conventional view of pruning as purely a compression
technique. DAP significantly outperforms domain-agnostic pruning
methods, is plug-and-play compatible with existing DG methods such as
CORAL and MIRO, and yields improved adversarial robustness by
eliminating exploitable domain-specific directions.

\paragraph{Limitations.}
DAP currently requires multiple source domains and performs
unstructured pruning, which does not yield wall-clock speedup without
sparse tensor support. Extending the domain sensitivity criterion to
structured pruning (filters or channels) is a natural next step that
would enable direct inference acceleration on commodity hardware.
By enabling efficient, generalizable models, DAP facilitates
deployment in resource-constrained and safety-sensitive settings such
as medical diagnosis and autonomous systems.



\newpage
\bibliographystyle{cas-model2-names}
\bibliography{cas-refs}
\appendix
\onecolumn

\section{Hyperparameter Configuration}
\label{appendix:hyperparameters}

Table~\ref{tab:hyperparameters} reports the full hyperparameter 
configuration used for DAP across all five benchmarks. All 
experiments use a ResNet-50 backbone pretrained with ERM, 
Adam optimizer for mask logits only, and Gumbel-Softmax 
relaxation with exponential temperature annealing from 
$\tau_\text{init}$ to $\tau_\text{final}$.

The majority of hyperparameters are shared across datasets, 
with three dataset-specific values: (i)~the domain sensitivity 
strength $\alpha$, which controls the scale of the gradient 
alignment signal relative to the learned logits; 
(ii)~the logit learning rate $\eta$; and 
(iii)~the sensitivity update schedule 
($f_\text{start}$, $f_\text{update}$).

The sensitivity strength $\alpha$ is the most critical 
hyperparameter. Because the domain sensitivity score 
$\bar{D} \in [-1, +1]$ (derived from sign agreement) is 
applied multiplicatively to logits via 
$\tilde{\lambda} = \lambda - \alpha \bar{D}$ 
(Eq.~11), $\alpha$ must be scaled commensurately with 
logit magnitudes to avoid either overwhelming the learned 
logits (if too large) or providing negligible signal 
(if too small). We find $\alpha \in \{0.5, 1.0, 2.0\}$.

\begin{table}[!htbp]
\centering
\caption{
    \textbf{Hyperparameter configuration for DAP.} 
    Shared parameters are identical across all datasets. 
    Dataset-specific values are listed in the lower section.
}
\label{tab:hyperparameters}
\resizebox{\columnwidth}{!}{%
\begin{tabular}{lll}
\toprule
\textbf{Category} & \textbf{Parameter} & \textbf{Value} \\
\midrule
\multicolumn{3}{l}{\textit{Shared across all datasets}} \\
\midrule
Architecture 
    & Backbone & ResNet-50 (ImageNet pretrained) \\
    & Pruning type & Unstructured (per-weight masks) \\
    & Weights & Frozen (mask logits only trained) \\
\midrule
Mask Learning 
    & Optimizer & Adam \\
    & Initial keep probability $p_\text{init}$ & 0.95 \\
    & Temperature $\tau_\text{init} \to \tau_\text{final}$ 
        & $2.0 \to 0.3$ (exponential) \\
    & Target sparsity $r_\text{target}$ & 0.999 \\
    & Total steps $T$ & 200 \\
\midrule
Adaptive $\lambda$ Scheduler 
    & Base $\lambda_\text{base}$ & 3.0 \\
    & Range $[\lambda_\text{min}, \lambda_\text{max}]$ & $[0.5, 50.0]$ \\
    & EMA smoothing $\beta_\lambda$ & 0.92 \\
    & Progress acceleration $a$ & 6.0 \\
    & Difficulty exponent $\rho$ & 1.2 \\
    & Target loss ratio $r_{\text{target}}$ & 0.1 \\
    & min loss ratio $r_{\text{min}}$ & 0.01 \\
    & max loss ratio $r_{\text{max}}$ & 1 \\
\midrule
Domain Sensitivity 
    & EMA smoothing $\beta_D$ & 0.92 \\
\midrule
Gradient Clipping 
    & Max norm & 4.0 \\
    & Max value & 6.0 \\
\midrule
\multicolumn{3}{l}{\textit{Dataset-specific}} \\
\midrule
\multicolumn{3}{l}{%
\begin{tabular}{@{}l r r r r r@{}}
\textbf{Parameter} & \textbf{PACS} & \textbf{VLCS} 
    & \textbf{OfficeHome} & \textbf{TerraInc} & \textbf{DomainNet} \\
\cmidrule(lr){1-1}\cmidrule(lr){2-6}
Sensitivity strength $\alpha$ 
    & 2.0      & 0.5      & 1.0    & 0.5      & 0.5      \\
Logit learning rate $\eta$ 
    & 0.002    & 0.0035   & 0.0035 & 0.002    & 0.0035   \\
Sensitivity start step $f_\text{start}$ 
    & 10{,}000 & 10{,}000 & 0      & 0        & 10{,}000 \\
Sensitivity update freq $f_\text{update}$ 
    & 30       & 1{,}000  & 500    & 1{,}000  & 1{,}000  \\
Batch size 
    & 32       & 32       & 32     & 32       & 16       \\
\end{tabular}%
}\\
\bottomrule
\end{tabular}%
}
\end{table}
\section{Per-Environment Results}
\label{appendix:per_environment}
Tables~\ref{tab:pacs_detailed}--\ref{tab:domainnet_detailed} report 
per-domain OOD accuracy for all five benchmarks at each sparsity 
level evaluated. Dense baseline results are taken from 
DomainBed~\citep{gulrajani2020search}; per-domain breakdowns for MIRO 
and SWAD are not publicly available, so only their published averages 
are included. For DAP, we additionally report the best checkpoint 
selected via source-domain validation (``Val.\ selection'') along 
with the sparsity at which it was selected. Throughout, \textbf{bold} 
marks the best result per domain and \underline{underline} marks the 
second best; $^\dagger$ denotes a sparse model.
\begin{table*}[!htbp]
\centering
\caption{\textbf{PACS} detailed results per domain.}
\label{tab:pacs_detailed}
\small
\resizebox{0.55\textwidth}{!}{%
\begin{tabular}{lccccc}
\toprule
\textbf{Method} & \textbf{Art} & \textbf{Cartoon} & \textbf{Photo} & \textbf{Sketch} & \textbf{Avg} \\
\midrule
\multicolumn{6}{l}{\textit{Dense Baselines}} \\
\midrule
ERM      & 85.6\;{\tiny$\pm$0.0} & 79.7\;{\tiny$\pm$0.0} & 97.9\;{\tiny$\pm$0.0} & 79.2\;{\tiny$\pm$0.0} & 85.7 \\
IRM      & 85.0\;{\tiny$\pm$1.6} & 77.6\;{\tiny$\pm$0.9} & 96.7\;{\tiny$\pm$0.3} & 78.5\;{\tiny$\pm$2.6} & 84.4 \\
GroupDRO  & 86.4\;{\tiny$\pm$0.3} & 79.9\;{\tiny$\pm$0.8} & 98.0\;{\tiny$\pm$0.3} & 72.1\;{\tiny$\pm$0.7} & 84.1 \\
Mixup    & 86.5\;{\tiny$\pm$0.4} & 76.6\;{\tiny$\pm$1.5} & 97.7\;{\tiny$\pm$0.2} & 76.5\;{\tiny$\pm$1.2} & 84.3 \\
MLDG     & 89.1\;{\tiny$\pm$0.9} & 78.8\;{\tiny$\pm$0.7} & 97.0\;{\tiny$\pm$0.9} & 74.4\;{\tiny$\pm$2.0} & 84.8 \\
CORAL    & 87.7\;{\tiny$\pm$0.6} & 79.2\;{\tiny$\pm$1.1} & 97.6\;{\tiny$\pm$0.0} & 79.4\;{\tiny$\pm$0.7} & 86.0 \\
MMD      & 84.5\;{\tiny$\pm$0.6} & 79.7\;{\tiny$\pm$0.7} & 97.5\;{\tiny$\pm$0.4} & 78.1\;{\tiny$\pm$1.3} & 85.0 \\
DANN     & 85.9\;{\tiny$\pm$0.5} & 79.9\;{\tiny$\pm$1.4} & 97.6\;{\tiny$\pm$0.2} & 75.2\;{\tiny$\pm$2.8} & 84.6 \\
MIRO     &  &  &  &  & 85.4 \\
SWAD     &  &  &  &  & \textbf{88.1} \\
\midrule
\multicolumn{6}{l}{\textit{DAP$^\dagger$ (Ours)}} \\
\midrule
20\%  & 84.9\;{\tiny$\pm$0.0} & 80.4\;{\tiny$\pm$0.2} & \underline{98.2\;{\tiny$\pm$0.0}} & 79.9\;{\tiny$\pm$0.1} & 85.8 \\
40\%  & \underline{86.7\;{\tiny$\pm$0.2}} & \underline{83.6\;{\tiny$\pm$0.0}} & \textbf{98.4\;{\tiny$\pm$0.1}} & \underline{82.3\;{\tiny$\pm$0.1}} & \underline{87.8} \\
60\%  & 85.7\;{\tiny$\pm$0.2} & \textbf{84.2\;{\tiny$\pm$0.3}} & 95.6\;{\tiny$\pm$0.2} & \textbf{83.3\;{\tiny$\pm$0.1}} & 87.2 \\
80\%  & 85.3\;{\tiny$\pm$0.3} & 83.2\;{\tiny$\pm$0.6} & 92.4\;{\tiny$\pm$0.3} & 83.2\;{\tiny$\pm$0.3} & 86.0 \\
\midrule
\multicolumn{6}{l}{\textit{DAP$^\dagger$ (Val.\ selection)}} \\
\midrule
Best  & 86.5\;{\tiny$\pm$0.2} & 83.8\;{\tiny$\pm$1.0} & 94.6\;{\tiny$\pm$0.6} & 83.3\;{\tiny$\pm$1.1} & 87.0 \\
{\tiny\textit{(Sparsity)}} & {\tiny\textit{(47\%)}} & {\tiny\textit{(65\%)}} & {\tiny\textit{(74\%)}} & {\tiny\textit{(67\%)}} & {\tiny\textit{(63\%)}} \\
\bottomrule
\end{tabular}%
}
\end{table*}

\begin{table*}[!htbp]
\centering
\caption{\textbf{VLCS} detailed results per domain.}
\label{tab:vlcs_detailed}
\small
\resizebox{0.55\textwidth}{!}{%
\begin{tabular}{lccccc}
\toprule
\textbf{Method} & \textbf{Caltech} & \textbf{LabelMe} & \textbf{SUN} & \textbf{VOC} & \textbf{Avg} \\
\midrule
\multicolumn{6}{l}{\textit{Dense Baselines}} \\
\midrule
ERM      & 98.6\;{\tiny$\pm$0.0} & 65.9\;{\tiny$\pm$0.1} & 68.9\;{\tiny$\pm$0.0} & 76.1\;{\tiny$\pm$0.0} & 77.4 \\
IRM      & 97.6\;{\tiny$\pm$0.3} & 65.0\;{\tiny$\pm$0.9} & 72.9\;{\tiny$\pm$0.5} & 76.9\;{\tiny$\pm$1.3} & 78.1 \\
GroupDRO  & 97.7\;{\tiny$\pm$0.4} & 62.5\;{\tiny$\pm$1.1} & 70.1\;{\tiny$\pm$0.7} & \underline{78.4\;{\tiny$\pm$0.9}} & 77.2 \\
Mixup    & 97.9\;{\tiny$\pm$0.3} & 64.5\;{\tiny$\pm$0.6} & 71.5\;{\tiny$\pm$0.9} & 76.9\;{\tiny$\pm$1.3} & 77.7 \\
MLDG     & 98.1\;{\tiny$\pm$0.3} & 63.0\;{\tiny$\pm$0.9} & \underline{73.5\;{\tiny$\pm$0.6}} & 73.7\;{\tiny$\pm$0.3} & 77.1 \\
CORAL    & \underline{98.8\;{\tiny$\pm$0.1}} & 64.6\;{\tiny$\pm$0.8} & 71.7\;{\tiny$\pm$1.4} & 75.8\;{\tiny$\pm$0.4} & 77.7 \\
MMD      & 97.1\;{\tiny$\pm$0.4} & 63.4\;{\tiny$\pm$0.7} & 71.4\;{\tiny$\pm$0.8} & 74.9\;{\tiny$\pm$2.5} & 76.7 \\
DANN     & 98.5\;{\tiny$\pm$0.2} & 64.9\;{\tiny$\pm$1.1} & 73.1\;{\tiny$\pm$0.7} & \textbf{78.3\;{\tiny$\pm$0.3}} & 78.7 \\
C-DANN   & 97.5\;{\tiny$\pm$0.1} & 65.2\;{\tiny$\pm$0.4} & 73.4\;{\tiny$\pm$1.1} & 76.9\;{\tiny$\pm$0.2} & 78.2 \\
MIRO     &  &  &  &  & \underline{79.0} \\
SWAD     &  &  &  &  & \textbf{79.1} \\
\midrule
\multicolumn{6}{l}{\textit{DAP$^\dagger$ (Ours)}} \\
\midrule
20\%  & 98.6\;{\tiny$\pm$0.0} & 64.5\;{\tiny$\pm$0.1} & \textbf{69.7\;{\tiny$\pm$0.2}} & 76.0\;{\tiny$\pm$0.1} & 77.2 \\
40\%  & \textbf{99.0\;{\tiny$\pm$0.1}} & 64.3\;{\tiny$\pm$0.2} & 66.8\;{\tiny$\pm$0.1} & 77.1\;{\tiny$\pm$0.3} & 76.8 \\
60\%  & 98.3\;{\tiny$\pm$0.3} & \textbf{68.4\;{\tiny$\pm$0.2}} & 62.4\;{\tiny$\pm$0.7} & 71.3\;{\tiny$\pm$0.3} & 75.1 \\
80\%  & 95.4\;{\tiny$\pm$1.5} & 63.0\;{\tiny$\pm$0.3} & 63.4\;{\tiny$\pm$0.5} & 69.5\;{\tiny$\pm$0.4} & 72.8 \\
\midrule
\multicolumn{6}{l}{\textit{DAP$^\dagger$ (Val.\ selection)}} \\
\midrule
Best  & 93.1\;{\tiny$\pm$3.4} & \underline{66.4\;{\tiny$\pm$0.8}} & 65.5\;{\tiny$\pm$0.5} & 69.5\;{\tiny$\pm$1.9} & 73.6 \\
{\tiny\textit{(Sparsity)}} & {\tiny\textit{(80\%)}} & {\tiny\textit{(73\%)}} & {\tiny\textit{(81\%)}} & {\tiny\textit{(83\%)}} & {\tiny\textit{(79\%)}} \\
\bottomrule
\end{tabular}%
}
\end{table*}
\begin{table*}[!htbp]
\centering
\caption{\textbf{OfficeHome} detailed results per domain.}
\label{tab:officehome_detailed}
\small
\resizebox{0.55\textwidth}{!}{%
\begin{tabular}{lccccc}
\toprule
\textbf{Method} & \textbf{Art} & \textbf{Clipart} & \textbf{Product} & \textbf{RealWorld} & \textbf{Avg} \\
\midrule
\multicolumn{6}{l}{\textit{Dense Baselines}} \\
\midrule
ERM      & 64.1\;{\tiny$\pm$0.0} & 52.4\;{\tiny$\pm$0.0} & 76.2\;{\tiny$\pm$0.0} & 78.2\;{\tiny$\pm$0.0} & 67.7 \\
IRM      & 61.8\;{\tiny$\pm$1.0} & 52.3\;{\tiny$\pm$1.0} & 75.2\;{\tiny$\pm$0.8} & 77.2\;{\tiny$\pm$1.1} & 66.6 \\
GroupDRO  & 61.6\;{\tiny$\pm$0.7} & 52.9\;{\tiny$\pm$0.2} & 75.5\;{\tiny$\pm$0.5} & 77.7\;{\tiny$\pm$0.2} & 66.9 \\
Mixup    & 64.7\;{\tiny$\pm$0.7} & 54.7\;{\tiny$\pm$0.6} & \underline{77.3\;{\tiny$\pm$0.3}} & 79.2\;{\tiny$\pm$0.3} & 69.0 \\
MLDG     & 63.7\;{\tiny$\pm$0.3} & 54.5\;{\tiny$\pm$0.6} & 75.9\;{\tiny$\pm$0.4} & 78.6\;{\tiny$\pm$0.1} & 68.2 \\
CORAL    & 64.4\;{\tiny$\pm$0.3} & 55.3\;{\tiny$\pm$0.5} & 76.7\;{\tiny$\pm$0.5} & 77.9\;{\tiny$\pm$0.5} & 68.6 \\
MMD      & 63.0\;{\tiny$\pm$0.1} & 53.7\;{\tiny$\pm$0.9} & 76.1\;{\tiny$\pm$0.3} & 78.1\;{\tiny$\pm$0.5} & 67.7 \\
DANN     & 59.3\;{\tiny$\pm$1.1} & 51.7\;{\tiny$\pm$0.2} & 74.1\;{\tiny$\pm$0.8} & 76.6\;{\tiny$\pm$0.6} & 65.4 \\
C-DANN   & 61.0\;{\tiny$\pm$1.4} & 51.1\;{\tiny$\pm$0.7} & 74.1\;{\tiny$\pm$0.3} & 76.0\;{\tiny$\pm$0.7} & 65.6 \\
MIRO     &  &  &  &  & \underline{70.5} \\
SWAD     &  &  &  &  & \textbf{70.6} \\
\midrule
\multicolumn{6}{l}{\textit{DAP$^\dagger$ (Ours)}} \\
\midrule
20\%  & \underline{65.4\;{\tiny$\pm$0.3}} & 54.1\;{\tiny$\pm$0.3} & \textbf{77.3\;{\tiny$\pm$0.2}} & \underline{79.2\;{\tiny$\pm$0.0}} & 69.0 \\
40\%  & \textbf{65.5\;{\tiny$\pm$0.1}} & \textbf{56.8\;{\tiny$\pm$0.2}} & 77.2\;{\tiny$\pm$0.2} & \textbf{80.2\;{\tiny$\pm$0.2}} & 69.9 \\
60\%  & 58.4\;{\tiny$\pm$1.1} & \underline{55.0\;{\tiny$\pm$0.4}} & 73.3\;{\tiny$\pm$0.3} & 76.7\;{\tiny$\pm$0.1} & 65.9 \\
80\%  & 54.9\;{\tiny$\pm$1.0} & 51.3\;{\tiny$\pm$0.6} & 69.7\;{\tiny$\pm$0.1} & 73.2\;{\tiny$\pm$0.0} & 62.3 \\
\midrule
\multicolumn{6}{l}{\textit{DAP$^\dagger$ (Val.\ selection)}} \\
\midrule
Best  & 56.5\;{\tiny$\pm$0.4} & 53.5\;{\tiny$\pm$0.5} & 71.2\;{\tiny$\pm$1.0} & 75.8\;{\tiny$\pm$0.7} & 64.2 \\
{\tiny\textit{(Sparsity)}} & {\tiny\textit{(80\%)}} & {\tiny\textit{(74\%)}} & {\tiny\textit{(74\%)}} & {\tiny\textit{(68\%)}} & {\tiny\textit{(74\%)}} \\
\bottomrule
\end{tabular}%
}
\end{table*}

\begin{table*}[!htbp]
\centering
\caption{\textbf{TerraIncognita} detailed results per domain.}
\label{tab:terra_detailed}
\small
\resizebox{0.55\textwidth}{!}{%
\begin{tabular}{lccccc}
\toprule
\textbf{Method} & \textbf{L100} & \textbf{L38} & \textbf{L43} & \textbf{L46} & \textbf{Avg} \\
\midrule
\multicolumn{6}{l}{\textit{Dense Baselines}} \\
\midrule
ERM      & 57.6\;{\tiny$\pm$0.0} & \underline{56.8\;{\tiny$\pm$0.0}} & 56.7\;{\tiny$\pm$0.0} & \underline{45.5\;{\tiny$\pm$0.0}} & 54.2 \\
IRM      & 52.2\;{\tiny$\pm$3.1} & 43.4\;{\tiny$\pm$2.4} & 57.7\;{\tiny$\pm$1.5} & 38.1\;{\tiny$\pm$0.7} & 47.9 \\
GroupDRO  & 47.2\;{\tiny$\pm$1.6} & 40.1\;{\tiny$\pm$1.6} & 57.6\;{\tiny$\pm$0.9} & 43.0\;{\tiny$\pm$0.7} & 47.0 \\
Mixup    & \underline{60.6\;{\tiny$\pm$1.3}} & 41.1\;{\tiny$\pm$1.8} & \underline{58.5\;{\tiny$\pm$0.8}} & 35.2\;{\tiny$\pm$1.1} & 48.9 \\
MLDG     & 48.5\;{\tiny$\pm$3.3} & 42.8\;{\tiny$\pm$0.4} & 56.8\;{\tiny$\pm$0.9} & 36.3\;{\tiny$\pm$0.5} & 46.1 \\
CORAL    & 48.6\;{\tiny$\pm$0.9} & 42.2\;{\tiny$\pm$3.5} & 55.9\;{\tiny$\pm$0.6} & 38.7\;{\tiny$\pm$0.7} & 46.4 \\
MMD      & 52.2\;{\tiny$\pm$5.8} & 47.0\;{\tiny$\pm$0.6} & \textbf{57.8\;{\tiny$\pm$1.3}} & 40.3\;{\tiny$\pm$0.5} & 49.3 \\
DANN     & 49.0\;{\tiny$\pm$3.8} & 46.3\;{\tiny$\pm$1.7} & 57.6\;{\tiny$\pm$0.8} & 40.6\;{\tiny$\pm$1.7} & 48.4 \\
C-DANN   & 49.5\;{\tiny$\pm$3.8} & 44.8\;{\tiny$\pm$1.0} & 57.3\;{\tiny$\pm$1.1} & 38.8\;{\tiny$\pm$1.7} & 47.6 \\
MIRO     &  &  &  &  & 50.4 \\
SWAD     &  &  &  &  & 50.0 \\
\midrule
\multicolumn{6}{l}{\textit{DAP$^\dagger$ (Ours)}} \\
\midrule
20\%  & 58.2\;{\tiny$\pm$0.1} & \textbf{54.7\;{\tiny$\pm$0.7}} & 56.2\;{\tiny$\pm$0.0} & \textbf{43.6\;{\tiny$\pm$0.1}} & \textbf{53.2} \\
40\%  & \textbf{58.7\;{\tiny$\pm$0.5}} & 49.2\;{\tiny$\pm$0.3} & 54.6\;{\tiny$\pm$0.0} & 41.9\;{\tiny$\pm$0.1} & \underline{51.1} \\
60\%  & 56.3\;{\tiny$\pm$0.9} & 41.0\;{\tiny$\pm$0.8} & 55.5\;{\tiny$\pm$0.1} & 40.0\;{\tiny$\pm$0.3} & 48.2 \\
80\%  & 57.8\;{\tiny$\pm$1.0} & 40.6\;{\tiny$\pm$1.8} & 54.4\;{\tiny$\pm$0.4} & 32.9\;{\tiny$\pm$0.5} & 46.4 \\
\midrule
\multicolumn{6}{l}{\textit{DAP$^\dagger$ (Val.\ selection)}} \\
\midrule
Best  & 57.2\;{\tiny$\pm$0.4} & 39.8\;{\tiny$\pm$2.0} & 55.0\;{\tiny$\pm$0.5} & 36.4\;{\tiny$\pm$0.4} & 47.1 \\
{\tiny\textit{(Sparsity)}} & {\tiny\textit{(72\%)}} & {\tiny\textit{(73\%)}} & {\tiny\textit{(76\%)}} & {\tiny\textit{(71\%)}} & {\tiny\textit{(73\%)}} \\
\bottomrule
\end{tabular}%
}
\end{table*}

\begin{table*}[!htbp]
\centering
\caption{\textbf{DomainNet} detailed results per domain.}
\label{tab:domainnet_detailed}
\small
\resizebox{0.7\textwidth}{!}{%
\begin{tabular}{lccccccc}
\toprule
\textbf{Method} & \textbf{Clipart} & \textbf{Infograph} & \textbf{Painting} & \textbf{Quickdraw} & \textbf{Real} & \textbf{Sketch} & \textbf{Avg} \\
\midrule
\multicolumn{8}{l}{\textit{Dense Baselines}} \\
\midrule
ERM      & 60.9\;{\tiny$\pm$0.0} & \underline{21.3\;{\tiny$\pm$0.0}} & 50.6\;{\tiny$\pm$0.0} & 13.2\;{\tiny$\pm$0.0} & 62.6\;{\tiny$\pm$0.0} & 53.3\;{\tiny$\pm$0.0} & 43.6 \\
IRM      & 51.0\;{\tiny$\pm$3.3} & 16.8\;{\tiny$\pm$1.0} & 38.8\;{\tiny$\pm$2.1} & 11.8\;{\tiny$\pm$0.5} & 51.5\;{\tiny$\pm$3.6} & 44.2\;{\tiny$\pm$3.1} & 35.7 \\
GroupDRO  & 47.8\;{\tiny$\pm$0.6} & 17.1\;{\tiny$\pm$0.6} & 36.6\;{\tiny$\pm$0.7} & 8.8\;{\tiny$\pm$0.4} & 51.5\;{\tiny$\pm$0.6} & 40.7\;{\tiny$\pm$0.3} & 33.7 \\
Mixup    & 55.3\;{\tiny$\pm$0.3} & 18.2\;{\tiny$\pm$0.3} & 45.0\;{\tiny$\pm$1.0} & 12.5\;{\tiny$\pm$0.3} & 57.1\;{\tiny$\pm$1.2} & 49.2\;{\tiny$\pm$0.3} & 39.6 \\
MLDG     & 59.5\;{\tiny$\pm$0.0} & 19.8\;{\tiny$\pm$0.4} & 48.3\;{\tiny$\pm$0.5} & 13.0\;{\tiny$\pm$0.4} & 59.5\;{\tiny$\pm$1.0} & 50.4\;{\tiny$\pm$0.7} & 41.8 \\
CORAL    & 58.7\;{\tiny$\pm$0.2} & 20.9\;{\tiny$\pm$0.3} & 47.3\;{\tiny$\pm$0.3} & \underline{13.6\;{\tiny$\pm$0.3}} & 60.2\;{\tiny$\pm$0.3} & 50.2\;{\tiny$\pm$0.6} & 41.8 \\
MMD      & 54.6\;{\tiny$\pm$1.7} & 19.3\;{\tiny$\pm$0.3} & 44.9\;{\tiny$\pm$1.1} & 11.4\;{\tiny$\pm$0.5} & 59.5\;{\tiny$\pm$0.2} & 47.0\;{\tiny$\pm$1.6} & 39.4 \\
DANN     & 53.8\;{\tiny$\pm$0.7} & 17.8\;{\tiny$\pm$0.3} & 43.5\;{\tiny$\pm$0.3} & 11.9\;{\tiny$\pm$0.5} & 56.4\;{\tiny$\pm$0.3} & 46.7\;{\tiny$\pm$0.5} & 38.4 \\
C-DANN   & 53.4\;{\tiny$\pm$0.4} & 18.3\;{\tiny$\pm$0.7} & 44.8\;{\tiny$\pm$0.3} & 12.9\;{\tiny$\pm$0.2} & 57.5\;{\tiny$\pm$0.4} & 46.7\;{\tiny$\pm$0.2} & 38.9 \\
MIRO     &  &  &  &  &  &  & 44.3 \\
SWAD     &  &  &  &  &  &  & \textbf{46.5} \\
\midrule
\multicolumn{8}{l}{\textit{DAP$^\dagger$ (Ours)}} \\
\midrule
20\%  & 62.2\;{\tiny$\pm$0.4} & \textbf{21.9\;{\tiny$\pm$0.2}} & \textbf{51.7\;{\tiny$\pm$0.4}} & 13.2\;{\tiny$\pm$0.4} & \textbf{63.3\;{\tiny$\pm$0.3}} & \textbf{54.8\;{\tiny$\pm$0.2}} & \underline{44.5} \\
40\%  & \underline{62.9\;{\tiny$\pm$0.3}} & 21.2\;{\tiny$\pm$0.1} & \underline{50.4\;{\tiny$\pm$0.6}} & \textbf{13.5\;{\tiny$\pm$0.5}} & \underline{61.8\;{\tiny$\pm$0.3}} & 53.2\;{\tiny$\pm$0.3} & 43.9 \\
60\%  & 62.5\;{\tiny$\pm$0.2} & 19.6\;{\tiny$\pm$0.1} & 48.8\;{\tiny$\pm$0.4} & 12.3\;{\tiny$\pm$0.5} & 59.8\;{\tiny$\pm$0.5} & 52.5\;{\tiny$\pm$0.4} & 42.6 \\
80\%  & \textbf{64.5\;{\tiny$\pm$0.3}} & 20.2\;{\tiny$\pm$0.3} & 50.3\;{\tiny$\pm$0.0} & 12.0\;{\tiny$\pm$0.4} & 60.3\;{\tiny$\pm$0.3} & \underline{53.8\;{\tiny$\pm$0.2}} & 43.5 \\
\midrule
\multicolumn{8}{l}{\textit{DAP$^\dagger$ (Val.\ selection)}} \\
\midrule
Best  & 64.9\;{\tiny$\pm$0.0} & 20.3\;{\tiny$\pm$0.2} & 50.6\;{\tiny$\pm$0.4} & 13.1\;{\tiny$\pm$0.3} & 60.8\;{\tiny$\pm$0.6} & 53.4\;{\tiny$\pm$0.4} & 43.9 \\
{\tiny\textit{(Sparsity)}} & {\tiny\textit{(76\%)}} & {\tiny\textit{(77\%)}} & {\tiny\textit{(77\%)}} & {\tiny\textit{(75\%)}} & {\tiny\textit{(76\%)}} & {\tiny\textit{(76\%)}} & {\tiny\textit{(76\%)}} \\
\bottomrule
\end{tabular}%
}
\end{table*}

\end{document}